\documentclass[10pt,twocolumn,letterpaper]{article}

\usepackage{iccv}
\usepackage{amsmath,amsfonts,bm}

\def\eqref#1{equation~\ref{#1}}
\def\1{\bm{1}}

\def\rvh{{\mathbf{h}}}

\def\rvp{{\mathbf{p}}}

\def\rvv{{\mathbf{v}}}
\def\rvw{{\mathbf{w}}}
\def\rvx{{\mathbf{x}}}
\def\rvy{{\mathbf{y}}}
\def\rvz{{\mathbf{z}}}

\def\mR{{\bm{R}}}
\def\mS{{\bm{S}}}

\DeclareMathAlphabet{\mathsfit}{\encodingdefault}{\sfdefault}{m}{sl}
\SetMathAlphabet{\mathsfit}{bold}{\encodingdefault}{\sfdefault}{bx}{n}

\usepackage{times}
\usepackage{epsfig}
\usepackage{graphicx}
\usepackage{amsmath, bm}
\usepackage{amssymb}
\usepackage{booktabs}
\usepackage{cite}
\usepackage{multirow}
\usepackage{subcaption}
\usepackage{color, xcolor}
\usepackage{colortbl}
\usepackage[accsupp]{axessibility}  % Improves PDF readability for those with disabilities

\definecolor{predcolor}{rgb}{0.74, 0.83, 0.9}

\newcommand{\singleappendix}[1]{%
  \appendix
  \section*{#1}
  \addcontentsline{toc}{section}{#1}
  \stepcounter{section}
}

\newcommand{\ourmethod}{NNCSL\xspace}
\newcommand{\changes}[1]{\textcolor{black}{#1}}

\usepackage[pagebackref=true,breaklinks=true,letterpaper=true,colorlinks,bookmarks=false]{hyperref}

\iccvfinalcopy % *** Uncomment this line for the final submission
\usepackage[capitalize]{cleveref}
\crefname{section}{Sec.}{Secs.}
\Crefname{section}{Section}{Sections}
\Crefname{table}{Table}{Tables}
\crefname{table}{Tab.}{Tabs.}

\ificcvfinal\fi

\begin{document}

%%%%%%%%% TITLE
\title{A soft nearest-neighbor framework for continual semi-supervised learning}

\author{Zhiqi Kang$^\ast$$^1$ \qquad Enrico Fini$^\ast$$^2$  \qquad Moin Nabi$^3$  \qquad Elisa Ricci$^{2,4}$ \qquad Karteek Alahari$^1$ \\\\ $^1$ Inria$^\dagger$ \qquad $^2$ University of Trento \qquad $^3$ SAP AI Research \qquad $^4$ Fondazione Bruno Kessler
% , Germany
}

\maketitle
\def\thefootnote{$\ast$}\footnotetext{\scriptsize{Zhiqi Kang and Enrico Fini contributed equally to this work}}
\def\thefootnote{$\dagger$}\footnotetext{\scriptsize{Univ.\ Grenoble Alpes, CNRS, Grenoble INP, LJK, 38000 Grenoble, France.}}
\ificcvfinal\thispagestyle{empty}\fi

%%%%%%%%% ABSTRACT
\begin{abstract}
Despite significant advances, the performance of state-of-the-art continual learning approaches hinges on the unrealistic scenario of fully labeled data. In this paper, we tackle this challenge and propose an approach for continual semi-supervised learning---a setting where not all the data samples are labeled. A primary issue in this scenario is the model forgetting representations of unlabeled data and overfitting the labeled samples. We leverage the power of nearest-neighbor classifiers to nonlinearly partition the feature space and flexibly model the underlying data distribution thanks to its non-parametric nature. This enables the model to learn a strong representation for the current task, and distill relevant information from previous tasks. We perform a thorough experimental evaluation and show that our method outperforms all the existing approaches by large margins, setting a solid state of the art on the continual semi-supervised learning paradigm. For example, on CIFAR-100 we surpass several others even when using at least 30 times less supervision (0.8\% vs.\ 25\% of annotations). Finally, our method works well on both low and high resolution images and scales seamlessly to more complex datasets such as ImageNet-100. Our source code is publicly available at \href{https://github.com/kangzhiq/NNCSL}{https://github.com/kangzhiq/NNCSL}.

\end{abstract}
\vspace{-0.2cm}
%%%%%%%%% BODY TEXT
\section{Introduction}
\vspace{-0.1cm}
\label{sec:intro}
Several efforts have been devoted to the continual learning (CL)~\cite{de2021continual} paradigm wherein training data samples arrive sequentially. 
However, most of the state-of-the-art CL methods~\cite{castro2018end, cha2021co2l, douillard2022dytox} are based on a strong assumption: the data is fully labeled. This is an unrealistic requirement as labeling data is oftentimes expensive for the expertise required or the amount of annotations, hazardous due to privacy or safety concerns, or impractical in a real-time online scenario. A natural way of tackling this issue is by leveraging the \textit{semi-supervised learning} framework, where not all the data samples are labeled.

\begin{figure}
\begin{subfigure}[h]{0.49\linewidth}
\includegraphics[width=\linewidth]{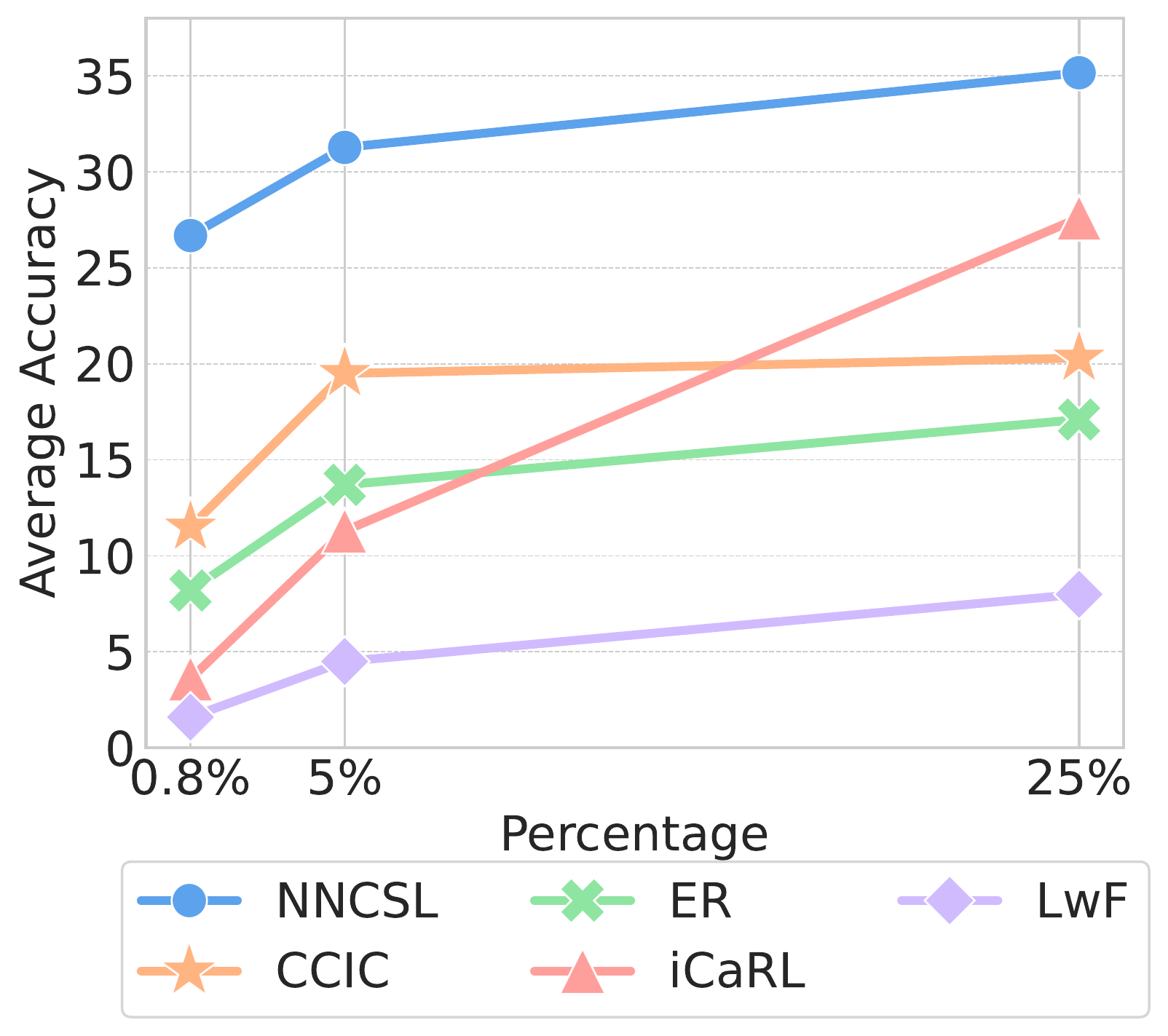}
% \caption{Training accuracy for labeled data}
\end{subfigure}
\begin{subfigure}[h]{0.49\linewidth}
\includegraphics[width=\linewidth]{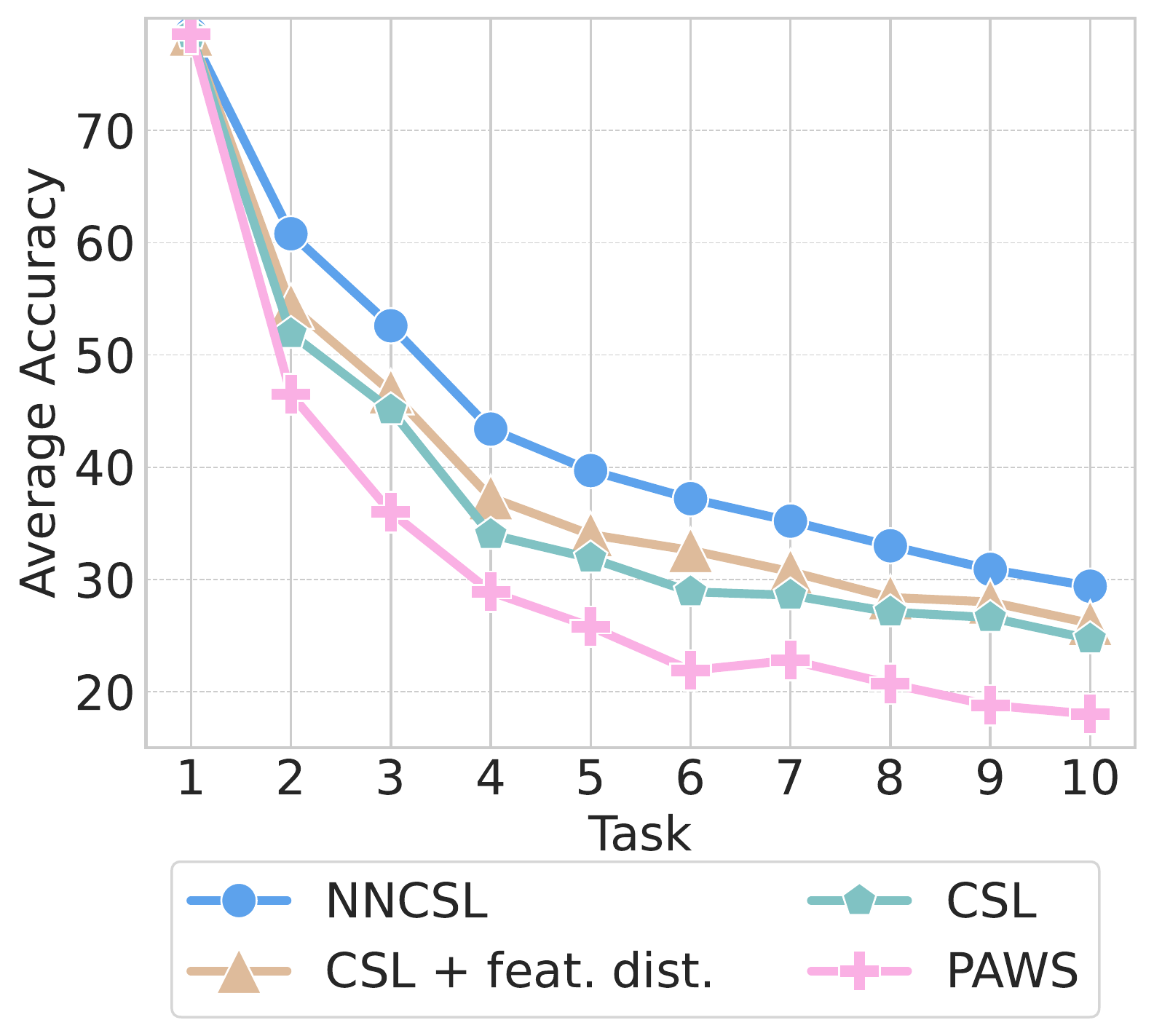}
% \caption{Training accuracy for unlabeled data}
\end{subfigure}
\vspace{-0.1cm}
\caption{Left: The average accuracy with different percentages of labeled data on CIFAR-100. Our method (NNCSL) with 0.8\% of the labels outperforms or matches the performance of all other methods at 25\%. Right: Comparison of different versions of our approach and PAWS \cite{assran2021semi}. CSL is equivalent to NNCSL without our NND loss.}
\label{fig:teaser}
\vspace{-0.5cm}
\end{figure}

In recent years, this stimulated the community to investigate a new line of research named \textit{continual semi-supervised learning} \cite{wang2021ordisco,boschini2022continual,smith2021memory}. It refers to the setting where each task in the sequence is semi-supervised.
This learning scenario brings novel challenges, as the models catastrophically forget the representations of unlabeled data while also overfitting the labeled set. This is further exacerbated by another well-studied phenomenon in CL: overfitting the experience replay buffer~\cite{buzzega2021rethinking}. These challenges result in vanilla CL methods underperforming, as they lack the ability to extract information from the unlabeled set, thus largely overfitting to the labeled set~\cite{wang2023comprehensive}.  On the other hand, semi-supervised learning approaches~\cite{lee2013pseudo,sohn2020fixmatch,chen2022debiased,zhai2019s4l} balance well the labeled and unlabeled sets but cannot handle the continual scenario, and suffer from forgetting even when paired with well-known CL methods (see Fig.~\ref{fig:teaser} (right) and Tab.~\ref{tab:cifar}).

A few recent approaches~\cite{wang2021ordisco,boschini2022continual} partially mitigate these issues on small-scale datasets.
% \changes{\cite{wang2021ordisco} relies on generative models to replay previous classes, which is not scalable, e.g., ImageNet-100, and leads to a sizeable computational and memory overhead. Another work~\cite{boschini2022continual} only samples positive and negative data at the task or class level, which is shown to be less effective with more classes, more samples, or higher resolution images (see Fig.~\ref{fig:teaser} (left) and results in Sec.~\ref{sec:experiments}).
% For instance, as the best-performing related approach, ~\cite{boschini2022continual} obtains only 32.3\% accuracy on ImageNet-100 (for reference,  our method reaches 65.3\% on the same setting). } 
% \changes{\cite{wang2021ordisco} relies on generative models to replay previous classes, which leads to a sizeable computational and memory overhead. For instance, their method is not applicable to large-scale datasets such as ImageNet-100. \cite{boschini2022continual} proposed to contrast samples of different classes and of different tasks. However, task-level contrastive learning showed little improvement in their ablation study. We speculate it may confuse the model as it gathers classes that are not necessarily related. This confusion leads to worse performance with less memory buffer, more samples, more tasks, or higher resolution images (see Fig.~\ref{fig:teaser} (left) and results in Sec.~\ref{sec:experiments}).}
\changes{The method in~\cite{wang2021ordisco} relies on generative models to replay previous classes, which leads to a sizeable computational and memory overhead, making it difficult to scale to larger datasets such as ImageNet. Boschini et al.~\cite{boschini2022continual} proposed to contrast among samples of different classes and tasks. However, \cite{boschini2022continual} falls short with smaller memory buffers, larger datasets, more tasks, or higher resolution images (see Fig.~\ref{fig:teaser} (left) and results in Sec.~\ref{sec:experiments}).}
%\cite{boschini2022continual} only samples positive and negative data at the task or class level, which is shown to be less effective with more classes, more samples, or higher resolution images (see Fig.~\ref{fig:teaser} (left) and results in Sec.~\ref{sec:experiments}).}
% For instance, ~\cite{boschini2022continual} obtains only 25.9\% accuracy on ImageNet-100 (for reference,  our method reaches 65.3\% on the same setting). 
% These approaches cannot learn effective representations from unlabeled data and their architecture is not scalable.
Therefore, we argue that there is a clear need for more powerful continual semi-supervised learning methods, with a more suitable  use of the labeled set, and efficient and stable representation learning from the unlabeled set.

In this paper, we unleash the power of nearest-neighbors in the context of continual semi-supervised learning. In particular, we propose a new method, NNCSL (\textbf{\underline{N}}earest-\textbf{\underline{N}}eighbor for \textbf{\underline{C}}ontinual \textbf{\underline{S}}emi-supervised \textbf{\underline{L}}earning), that leverages the ability of the nearest-neighbor classifier to non-linearly partition the feature space in two ways: i) to learn powerful and stable representations of the current task using a self-supervised multi-view strategy, and ii) to distill previous knowledge and transfer the local structure of the feature space. The 
latter is achieved through our proposed NND (\textbf{\underline{N}}earest-\textbf{\underline{N}}eighbor \textbf{\underline{D}}istillation), a novel semi-supervised distillation loss that mitigates forgetting in continual semi-supervised learning better than other competitive distillation approaches. In contrast with knowledge distillation~\cite{li2018pami, castro2018end, rebuffi2017cvpr,fini2020online} and feature distillation~\cite{hou2019learning, douillard2020podnet,fini2022self}, which focus exclusively on class-level and sample-level distributions respectively, NND simultaneously distills relationships between classes and samples by leveraging the nearest-neighbor classifier. Overall, NNCSL outperforms all related methods by very large margins on both small and large scale datasets and both low and high resolution images. For instance, as shown in Fig.~\ref{fig:teaser}, NNCSL matches or surpasses all others with more than 30 times less supervision (0.8\% vs.\ 25\% of annotations) on CIFAR100.

The main \textbf{contributions} of this work are as follows:
\begin{itemize}
    \item We propose NNCSL, a novel nearest-neighbor-based continual semi-supervised learning method that is, by design, impacted less by the overfitting phenomenon related to a small labeled buffer.
    \item We propose NND, a new distillation strategy that transfers both representation-level and class-level knowledge from the previously trained model using the outputs of a soft nearest-neighbor classifier, which effectively helps alleviate forgetting.
    \item We show that NNCSL outperforms the existing methods on several benchmarks by large margins, setting a new state of the art on continual semi-supervised learning. In contrast to previous approaches, our method works well on both low and high resolution images and scales seamlessly to more complex datasets.
\end{itemize}

\section{Related work}
\vspace{-0.1cm}
\label{sec:related_work}
\noindent\textbf{Semi-supervised learning.}
Semi-supervised methods focus on learning models from large-scale datasets where only a few samples have associated annotations \cite{chen2022semi}. 
% Since deep networks have become the mainstream in semi-supervised learning, several different approaches  have been introduced.
Early strategies for this learning paradigm applied to deep architectures leveraged pseudo-labels and performed self-training based on them \cite{lee2013pseudo}. This scheme was later improved with confidence thresholding \cite{arazo2020pseudo} and adaptive confidence thresholding~\cite{xu2021dash, zhang2021flexmatch}. More sophisticated methods for incorporating the confidence of the predictions and filtering out spurious samples were also developed, such as FixMatch~\cite{sohn2020fixmatch}, which employs a student-teacher architecture. 
Other approaches demonstrated the benefit of co-training~\cite{qiao2018deep} and  distillation~\cite{xie2020self}.

Another class of approaches was derived with the idea of imposing similar predictions from the network for two samples obtained with different input perturbations~\cite{sajjadi2016regularization, laine2017temporal, miyato2016distributional, assran2021semi, park2018adversarial, tarvainen2017mean, athiwaratkun2019there, zhang2020wcp, verma2019interpolation}. For example, \cite{assran2021semi} considered a consistency loss and soft pseudo labels generated by comparing the representations of the image views to those of a set of randomly-sampled labeled images.
% Different strategies can be also combined and there exist methods which integrate both pseudo-labeling and consistency regularization. 
Recently, sample mixing techniques, such as MixUp, were also investigated in the context of semi-supervised learning for improving the model performance on low sample density regions \cite{berthelot2019mixmatch, berthelot2020remixmatch, liu2022decoupled}. 
% Ideas from self-supervised methods 
% were also introduced into semi-supervised learning. For instance, self-supervised pre-training was found beneficial in~\cite{cai2021exponential}, exponential moving average normalization was adopted in~\cite{cai2021exponential} and contrastive learning was considered in~\cite{assran2020supervision}. 
However, none of the aforementioned works addressed the problem of learning in an incremental setting.

\noindent\textbf{Continual learning.} Several CL approaches have been proposed in the last few years to learn from data in an incremental fashion. According to a recent survey~\cite{de2021continual}, existing CL methods can be roughly categorized into three groups. The first category comprises regularization-based methods, which address the problem of catastrophic forgetting by introducing appropriate regularization terms in the objective function~\cite{cha2021co2l, douillard2020podnet, fini2020online, hou2019learning, li2018pami} or identifying a set of parameters that are most relevant for certain tasks~\cite{chaudhry2018riemannian, kirkpatrick2017overcoming, wu2019large}.
Replay-based methods correspond to the second group, and they store a few samples from previous tasks~\cite{buzzega2020dark,  Chaudhry19,Lopez-Paz17, rebuffi2017cvpr} or generate them~\cite{shin2017continual,ostapenko2019learning } in order to rehearse knowledge during the training phases for subsequent tasks.
Finally, the third category is parameter isolation
methods~\cite{Rusu16progressive, serra2018overcoming}, which operate by allocating task-specific parameters.

While the vast majority of these methods operate in a supervised setting, recent works addressed the problem of overcoming catastrophic forgetting in the challenging case of limited   \cite{ lechat2021semi, smith2021memory,boschini2022continual, wang2021ordisco} or no supervision~\cite{fini2022self, rao2019continual, smith2019unsupervised, achille2018life}. However, most of them have default settings that are significantly different, e.g., the use of external datasets, and the accessibility of labeled/unlabeled data during continual learning stages, leaving only a few~\cite{wang2021ordisco,boschini2022continual} to be comparable in our desired realistic setting. 
Wang {et al.}~\cite{wang2021ordisco} addressed the continual semi-supervised learning problem and proposed ORDisCo,
a method that continually learns a conditional generative adversarial network with a
classifier from partially labeled data. \changes{However, the method yields prohibitive costs on higher resolution images, e.g., on ImageNet-100.} Contrastive continual interpolation consistency (CCIC)
~\cite{boschini2022continual} \changes{is another approach, which proposed to contrast samples among different classes and tasks. However, task-level contrastive learning showed insignificant improvement in the ablation study. We speculate it may confuse the model as it gathers classes that are not necessarily related in the feature   space. This confusion leads to worse performance with a small memory buffer, more samples, more tasks, or higher resolution images as shown by our experiments.}
% ~\cite{boschini2022continual} is another approach, which leverages metric learning and consistency regularization for extracting knowledge from unlabeled samples. 
Our work radically departs from these previous methods, as we design \ourmethod, a novel approach for continual semi-supervised learning based on a soft nearest-neighbor classifier. Our empirical evaluation demonstrates that \ourmethod surpasses prior works by a large margin.

\section{Continual semi-supervised learning}
\begin{figure}
\includegraphics[width=\linewidth]{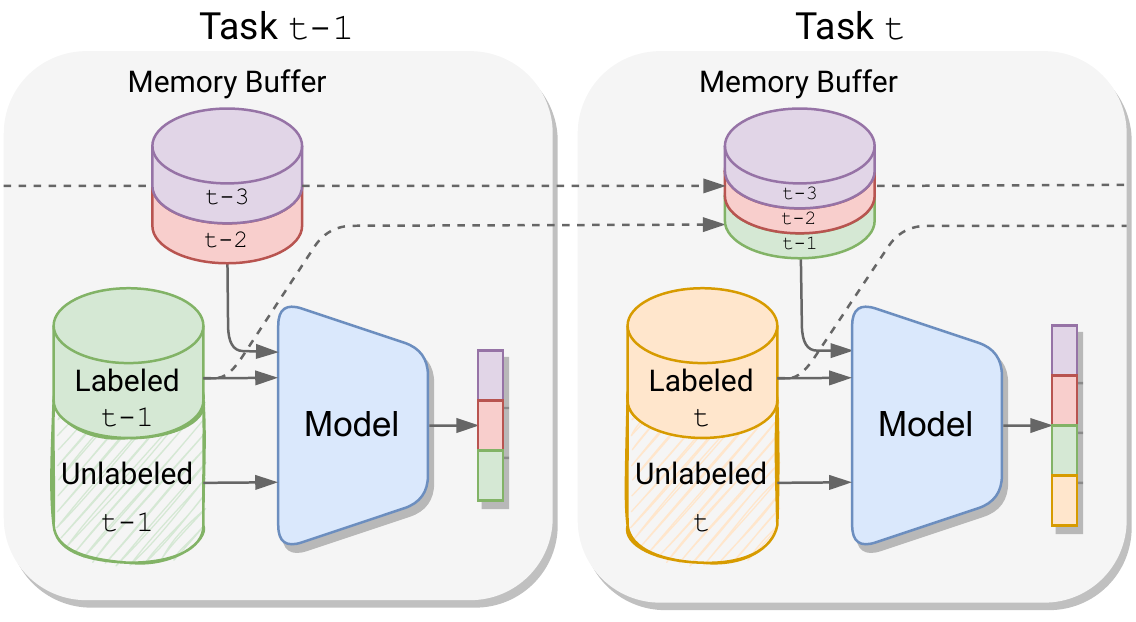}
\vspace{-0.7cm}
% \caption{Training accuracy for labeled data}
\caption{Illustration of the learning process in continual semi-supervised learning.}
\label{fig:continual-semi-sup}
\vspace{-0.3cm}
\end{figure}

\vspace{-0.1cm}
\label{sec:semi_CL}
We now formally define the problem of continual semi-supervised learning. Let the training data arrive sequentially, i.e., as a sequence of $T$ tasks. The dataset associated to task $t$ is denoted as $D_t$, with $t \in \{1, ..., T\}$. Learning is therefore performed task-wise, where only the current training data $D_t$ is available during task $t$. When switching from one task to the next one, previous data is systematically discarded. Since the available dataset is not fully labeled, we further divide it into two subsets such that $D_t = U_t \cup L_t $. 
Typically in a semi-supervised learning scenario, we have $|L_t| \ll |U_t|$, the ratio $|L_t| / |U_t|$ is kept constant for all the tasks. In addition, it is common practice in the CL literature \cite{rebuffi2017cvpr,boschini2022continual} to allow the retention of a memory buffer $M$ that stores and replays previously seen samples, as shown in Fig. \ref{fig:continual-semi-sup}.

Let $f_{\theta}$ be the model, parameterised by $\theta$, and consisting of three components: a backbone $g$, a projector $h$ and a classifier $p$. The backbone, here modeled as a convolutional neural network, is used to extract representations $\rvz = g(\rvx)$ from an input image $\rvx$. The classifier takes this representation to predict a set of logits $\rvp = p(\rvz)$, while the projector (implemented as a multi-layer perceptron) maps the backbone features to a lower-dimensional space $\rvh = h(\rvz)$. In addition, we use superscript to refer to the state at a certain point in time, for instance for task $t$ as $f_\theta^t$, and for the previous task $t-1$ as $f_\theta^{t-1}$. Similarly, we use $\rvx_u^t$ and $\rvx_l^t$ to refer to samples drawn from $U_t$ and $L_t$ respectively. Apart from images, the labeled dataset also contains one-hot ground truth annotations $\rvy$.

In the following sections, we introduce the proposed \ourmethod method for continual semi-supervised learning. We first present PAWS~\cite{assran2021semi}, which inspired our \ourmethod, (Sec.~\ref{subsec:paws}) and show why this method is not immediately applicable to the continual setting. Subsequently, we present CSL, our base continual semi-supervised learner  (Sec.~\ref{subsec:filtering}), which solves many of its issues. However, this base method lacks a mechanism to counteract forgetting. Hence, we introduce NND, our novel distillation approach based on the soft nearest-neighbor classifier in Sec.~\ref{subsec:distillation}. All these elements are harmoniously integrated into in our full method: NNCSL, whose overall objective is summarized in Sec.~\ref{subsec:overall_obj}.

\section{Nearest-neighbor meets continual learning: strengths and weaknesses}
\vspace{-0.1cm}
\label{sec:method}
\label{subsec:paws}
We now discuss the use of nearest-neighbor techniques~\cite{assran2021semi} in the context of continual learning, and describe its strengths and weaknesses in scenarios with data distribution shifts. During training, the mini-batches that the model receives are composed of labeled and unlabeled data, with $K$ and $N$ as batch sizes for these two sets respectively. Unlabeled images in the batch are augmented twice using common data augmentation techniques to obtain two correlated views of the same sample $(\rvx, \hat{\rvx})$. The model processes the batch, producing the projected representations $\rvh_l$ and ($\rvh_u, \hat{\rvh}_u)$ for labeled and unlabeled samples respectively.

The main idea from~\cite{assran2021semi} is to assign pseudo-labels for unlabeled samples in a non-parametric manner by considering their relationship with labeled samples, i.e., nearest-neighbor label assignments. Samples are compared in the feature space using the cosine similarity of projected features, and then the pseudo label is obtained by aggregating labels according to the similarities. More formally, let the superscript $k$ represent the index of the $k^{th}$ sample in the labeled mini-batch, and $\delta$ be the cosine similarity. One can apply a soft nearest-neighbor classifier to classify the augmented unlabeled sample $\hat{\rvx}_u$ as follows:
%calculate the distance, or inversely the similarity, between two samples by using their extracted representation with $d(a, b)$. Hence, we can apply a $K$ nearest neighbor between one unlabeled sample and $K$ labeled samples to classify the unlabeled one, with $K$ the number of labeled data in one mini-batch. At train time, we assign a soft nearest neighbor label $y^u$ to the unlabeled sample $x^u $ as: 
\begin{equation}
\label{eq:snn}
    \hat{\rvv} = \operatorname{SNN}(\hat{\rvh}_u, \mS, \epsilon) = \sum_{k}^{K} \frac{e^{c_k\delta(\hat{\rvh}_u, \rvh^k_l) / \epsilon}}{ \sum_{i}^{K} e^{c_i\delta(\hat{\rvh}_u, \rvh^i_l)/\epsilon}} \, \rvy^k,
\end{equation}
where $\mS = [\rvh^1_l,...,\rvh^K_l]$ are the features of the support samples and $\epsilon$ is a sharpening parameter that controls the entropy of the pseudo-label. Similarly, we can classify the other view of the same sample $\rvv = \operatorname{SNN}(\rvh_u,  \mS, \tau)$, with the only difference that we use a more gentle sharpening parameter $\tau > \epsilon$, referred to as the temperature.  Now, we can use $\hat{\rvv}$ as a target pseudo-label and train the network through the cross-entropy loss:
\begin{equation}
    \mathcal{L}_{\operatorname{SNN}} = H(\rvv, \hat{\rvv}).
\end{equation}
% By nature, this loss is asymmetric, but it can be symmetrized by swapping the two views.

The mechanism described above encourages the network to output consistent representations for similar inputs, while also accounting for the distribution of the classes in the feature space. However, one issue with this formulation is that the network could output unbalanced or even degenerate predictions where some classes are predicted more frequently than others. To avoid this, PAWS imposes the distribution-wise likelihood of all the classes to be uniform using a regularization term called Mean Entropy Maximization (MEM) loss defined as:\footnote{With a slight abuse of notation we refer to $H(\cdot)$ with one argument as the entropy function, while when two arguments are passed we consider it as the cross-entropy function $H(\cdot, \cdot)$.}
\begin{equation}
    \mathcal{L}_{\operatorname{MEM}} = H\left(\frac{1}{N}\sum^{N}_{n} \hat{\rvv}_n\right).
\end{equation}
Given these two losses, the total loss for PAWS is a weighted average of the two.

The advantage of this soft nearest-neighbor formulation is that it utilizes labeled samples as support vectors, not as training samples, which reduces overfitting. 
This property is interesting from the point of view of continual learning, since we would like to extract as much training signal as possible from the memory buffer without overfitting to it. However, PAWS is not designed to work under data distribution shifts. The key issue of PAWS in the CL setting is the assumption that the labeled and unlabeled sets exhibit the same distribution. This is untenable in CL, as the memory buffer contains classes not in the current task's unlabeled set. MEM loss aggravates this problem, as it tries to scatter the pseudo label over all the classes, even for the ones whose unlabeled samples are unavailable. A simple solution would be to use the labeled data of the current task and discard the buffer, but this is sub-optimal as the buffer is critical for CL.

\changes{Another important drawback of PAWS in the context of CL is the ambiguity of two-stage methods. PAWS performs best when the learned backbone is fine-tuned using the labeled set. This two-stage pipeline (pre-training and then fine-tuning) is prone to a loss in generalization performance as the labeled set is very small. \textit{Offline} methods such as PAWS can afford this loss to gain specialized knowledge on the targeted tasks. However, CL requires the reuse of the model for the next task. On one hand, using the fine-tuned model leads to a noticeable loss in performance on the subsequent tasks. On the other hand, using the pre-trained checkpoint (before fine-tuning) preserves more general properties but brings additional memory overhead, as two models need to be stored, which is undesirable in CL. In the following section, we describe our solution to overcome these limitations. A detailed illustration of the ambiguity of two-stage methods can be found in the supplementary material.
} 

% Another important drawback of PAWS in the context of CL is that it performs best when the learned backbone is fine-tuned using the linear classifier together with the labeled set. This two-stage pipeline (pre-training and then fine-tuning) is prone to a loss of generalizability as the labeled set is very small. \textit{Offline} methods such as PAWS can afford this to gain specialized knowledge on the targeted tasks and computational efficiency (linear classifier vs. nearest-neighbor) at inference time. However, in CL it is unclear which checkpoint of the two-stage methods should be used to further train the model on subsequent tasks. Using the checkpoint after pre-training preserves a more general model, but brings additional memory overhead by storing models from both stages, which is undesirable in a continual setting. In contrast, using the checkpoint after the fine-tuning results in a noticeable loss of performance on the subsequent tasks. In the following section, we describe our solution to overcome these limitations.

\begin{figure}
    \centering
    \includegraphics[width=\linewidth]{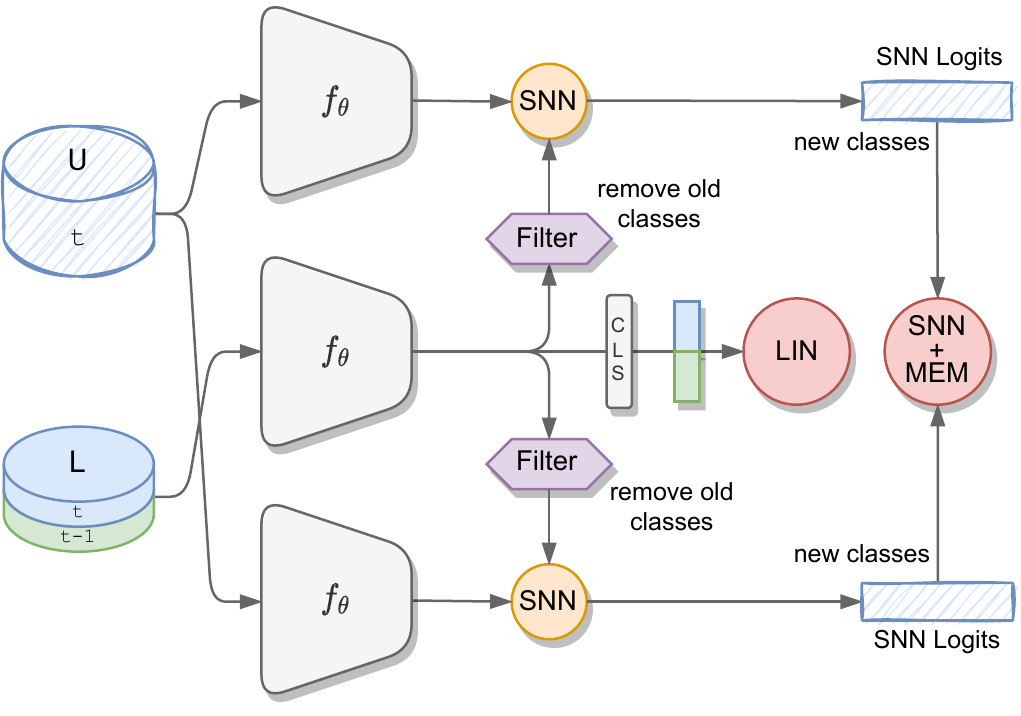}
    \vspace{-0.7cm}
    \caption{Overview of the base learner component of our method, which does not have a distillation loss. We refer to this as CSL.}
    \label{fig:overall}
    \vspace{-0.3cm}
\end{figure}

\section{NNCSL: Our nearest-neighbor approach for continual semi-supervised learning}
\label{sec:NNCSL}
\subsection{Our continual semi-supervised learner}
\label{subsec:filtering}
We now describe our proposed approach that leverages the strengths of the nearest-neighbor approach described in Sec.~\ref{subsec:paws}, while overcoming its weaknesses. The easiest way to make the labeled and unlabeled distributions match is to disregard the memory buffer. This is clearly undesirable for CL. However, one could multi-task PAWS with another objective that also takes into account the information of the memory buffer. In particular, we suggest processing labeled samples of all the classes seen so far, but filtering out samples from the previous tasks so they do not interfere in the computation of Eq.~\ref{eq:snn}. However, we can use the output of the linear classifier $p$ and optimize a standard cross-entropy loss:
\begin{equation}
\vspace{-0.05cm}
    \mathcal{L}_{\operatorname{LIN}} = \sum_j^J \operatorname{H}(\rvp^k, \rvy^j),
\vspace{-0.05cm}
\end{equation}
on all the $J$ labeled samples in the current batch (which also contains $K$ labeled samples of the current task).
The complete loss for our base continual semi-supervised learner (named CSL) is as follows:
\begin{equation}
\vspace{-0.05cm}
    \mathcal{L}_{\operatorname{CSL}} = 
    \mathcal{L}_{SNN} + \lambda_{\operatorname{MEM}} \cdot \mathcal{L}_{\operatorname{MEM}} + \lambda_{\operatorname{LIN}} \cdot \mathcal{L}_{\operatorname{LIN}}.
\vspace{-0.05cm}
\end{equation}
This loss has several favorable effects; it: i) stimulates the network to focus on the old classes while learning representations of the new ones through PAWS, ii) creates an ensemble effect between the two classifiers, iii) completely removes the need for fine-tuning, as the linear classifier is trained online, and iv) enables us to control the trade-off between fitting labeled or unlabeled data through the parameter $\lambda_{\operatorname{LIN}}$. Interestingly, we found that very small values of $\lambda_{\operatorname{LIN}}$ work well in practice, while larger values increase overfitting. 
We believe that, due to its partially self-supervised nature, $\mathcal{L}_{\operatorname{SNN}}$ learns improved representations, that can be easily discriminated by the linear classifier.
An illustration of the architecture of CSL is shown in Fig.~\ref{fig:overall}.

\begin{figure}
    \centering
    \includegraphics[width=\linewidth]{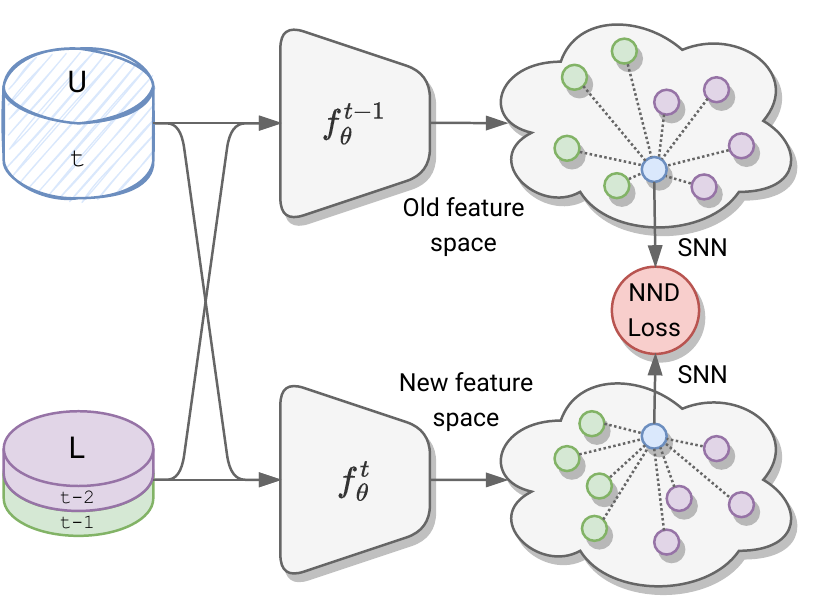}
    \vspace{-0.7cm}
    \caption{Illustration of our proposed NND loss, which compares the predictions of $f_{\theta}^{t} $ and $f_{\theta}^{t-1}$ with the same unlabeled samples and support samples.
    }
    \label{fig:distillation}
    \vspace{-0.3cm}
\end{figure}

\subsection{Soft nearest-neighbor distillation}
\label{subsec:distillation}
Distilling information~\cite{hinton2015distilling} is a common practice in CL, which utilizes frozen models (or a bank of features and/or probabilities) trained on previous tasks as a teacher to regularize the currently active model, which is a student. Let $t$ be the index for the current task. The student model $f_{\theta}^{t} $ aims to mimic the outputs of the teacher $f_{\theta}^{t-1}$, while learning the new task. Previous works~\cite{li2018pami,fini2020online} typically distill either the logits of a linear classifier or the features of the hidden layers of the network. However, in CSL, the main driver for the network to learn representations is the loss applied to the soft nearest-neighbor classifier. As explained in Sec.~\ref{subsec:filtering}, this loss does not give any signal on previous data, as old samples get filtered out and fed to the linear classifier only. This is made worse by the fact that the nearest-neighbor classifier is applied on a separated projection head $h$, that has no incentive to remember previous knowledge. 

To mitigate these issues, we devise a novel Nearest-Neighbor Distillation (NND) loss that blends well with our framework. The loss is based on the intuition that we can evaluate the nearest-neighbor classifier on the old feature space using the same support samples. This equates to computing the following two vectors: $\rvw = \operatorname{SNN}(\rvh_u, \mR, \tau)$ and $\rvw^{t-1} = \operatorname{SNN}(\rvh^{t-1}_u, \mR^{t-1}, \tau)$, where $\rvh^{t-1}_u$ is a feature vector output by the teacher for an unlabeled sample $\rvx_u$, while $\mR$ and $\mR^{t-1}$ represent the support set of previous classes embedded in the old and new feature spaces respectively. To mitigate forgetting, we use the probabilities predicted by the teacher as a distillation target:
\begin{equation}
\label{eq:snn_dist_loss}
    \mathcal{L}_{\operatorname{NND}} = H(\rvw, \rvw^{t-1}).
\end{equation}
Note that the output of the teacher is not sharpened as it is done in Eq.~\ref{eq:snn}. We apply the same temperature for both new and old features. We emphasize that here we use an inverted filter as that of Sec. \ref{subsec:filtering}, to distill knowledge about the previous classes only. See Fig.~\ref{fig:distillation} for a visual intuition and supplementary material for the visualization of its impact on deep features. It is worth mentioning that the memory overhead introduced by storing $f_{\theta}^{t-1}$ can be easily reduced by storing instead the features, which are lightweight compared to images. For instance, storing a 128-dimension feature takes roughly
$1/1000^{th}$ memory w.r.t. an RGB image of size 224 × 224. 

Our NND loss is different from the standard distillation loss. Knowledge distillation focuses on class-level distributions calculated by comparing samples of the new tasks with the prototype of each class. This causes a loss of information about representations of individual samples in the latent space. On the contrary, NND distills the aggregated relationships of each unlabeled sample with respect to all the labeled ones in the mini-batch. This encourages the model to maintain more stable representations, by anchoring unlabeled samples to labeled samples. In addition, NND can capture and transfer non-linear sample-class relationships by definition, unlike knowledge distillation which is limited to linear boundaries. Also, the $\operatorname{SNN}$ classifier is computed on a different support set sampled from the buffer at every iteration, which introduces randomness that further regularizes the model.  Furthermore, when distilling at very low temperatures (e.g., 0.1), the \textit{softmax} function from Eq.~\ref{eq:snn} behaves like an \textit{argmax} (i.e., selecting the closest samples in the feature space with high cosine similarity), leading to the transfer of information mainly on the local neighborhood from the teacher to the student. Finally, with respect to feature distillation which lacks alignment with class labels, NND carries more information about class distributions, which results in improved performance.

\subsection{Overall loss}
\label{subsec:overall_obj}
The overall NNCLS model, composed of two novel components, \ie, CSL and NND, is trained with the loss:
\begin{equation}
    \mathcal{L}_{\operatorname{NNCSL}} = \mathcal{L}_{\operatorname{CSL}} + \lambda_{\operatorname{NND}} \cdot \mathcal{L}_{\operatorname{NND}}.
\end{equation}

\section{Evaluation and analysis}
\vspace{-0.1cm}
\label{sec:experiments}
\begin{table*}[h]
  \centering
  \begin{tabular}{l*6c}
  \toprule
    % \textbf{Dataset} & \multicolumn{3}{c}{CIFAR-10(92.12$\pm$0.1)} & \multicolumn{3}{c}{CIFAR-100(67.7$\pm$0.9)}   \\
    \multirow{2}[1]{*}{\textbf{Method}} & \multicolumn{3}{c}{\textbf{CIFAR-10}} & \multicolumn{3}{c}{\textbf{CIFAR-100}}   \\
\cmidrule(lr){2-4}
\cmidrule(lr){5-7}
     & 0.8\% & 5\% & 25\%   & 0.8\% & 5\% & 25\% \\
    \midrule
    Fine-tuning & 13.6$\pm$2.9 & 18.2$\pm$0.4 & 19.2$\pm$2.2 & 1.8$\pm$0.2 & 5.0$\pm$0.3 & 7.8$\pm$0.1  \\
    % LwF  \cite{li2018pami}       & 13.1$\pm$2.9 & 17.7$\pm$3.2 & 19.4$\pm$1.7 & 1.6$\pm$0.1 & 4.5$\pm$0.1 & 8.0$\pm$0.1  \\
    oEWC \cite{kirkpatrick2017overcoming}        & 13.7$\pm$1.2 & 17.6$\pm$1.2 & 19.1$\pm$0.8 & 1.4$\pm$0.1 & 4.7$\pm$0.1 & 7.8$\pm$0.4  \\
    \midrule 
    ER (500)  \cite{rolnick2019experience}        & 36.3$\pm$1.1 & 51.9$\pm$4.5 & 60.9$\pm$5.7 & 8.2$\pm$0.1 & 13.7$\pm$0.6& 17.1$\pm$0.7\\
    iCaRL (500) \cite{rebuffi2017cvpr}     & 24.7$\pm$2.3 & 35.8$\pm$3.2 & 51.4$\pm$8.4 & 3.6$\pm$0.1 & 11.3$\pm$0.3& 27.6$\pm$0.4\\
    FOSTER (500) \cite{wang2022foster}      & 43.3$\pm$0.7 & 51.9$\pm$1.3 & 57.1$\pm$2.0 & 4.7$\pm$0.6  & 14.1$\pm$0.6  & 21.7$\pm$0.7\\
    X-DER (500) \cite{boschini2022class}      &  33.4$\pm$1.2 & 48.2$\pm$1.7 & 58.9$\pm$1.5 & 8.9$\pm$0.3  & 18.3$\pm$0.5 & 23.9$\pm$0.7\\
    % GDumb (500) \cite{prabhu20}      & 39.6$\pm$9.6 & 40.9$\pm$11.8& 44.8$\pm$5.4 & 8.6$\pm$0.1 & 9.9$\pm$0.4 & 10.1$\pm$0.4\\
    \midrule 
    PseudoER (500)    & 50.5$\pm$0.1& 56.5$\pm$0.6 & 57.0$\pm$0.6 &    8.7$\pm$0.4   & 11.4$\pm$0.5 & 12.3$\pm$0.2\\
    CCIC \cite{boschini2022continual} (500)        & 54.0$\pm$0.2 & 63.3$\pm$1.9 & 63.9$\pm$2.6 &               11.5$\pm$0.7& 19.5$\pm$0.2& 20.3$\pm$0.3\\
    PAWS \cite{assran2021semi} (500)        & 51.8$\pm$1.6 & 64.6$\pm$0.6& 65.9$\pm$0.3 & 16.1$\pm$0.4 & 21.2$\pm$0.4 &  19.2$\pm$ 0.4\\
    CSL (500)        & 64.5$\pm$0.7  & 69.6$\pm$0.5 & 70.0$\pm$0.4 & 23.6$\pm$0.3 & 26.2$\pm$0.5 &  29.3$\pm$0.3\\
    \rowcolor{predcolor} \ourmethod (500)        & \textbf{73.2}$\pm$0.1&  \textbf{77.2}$\pm$0.2 &  \textbf{77.3}$\pm$0.1 & \textbf{27.4}$\pm$0.5 & \textbf{31.4}$\pm$0.4 & \textbf{35.3}$\pm$0.3 \\
    \midrule
    PseudoER (5120)         &  55.4$\pm$0.5 & 70.0$\pm$0.3 & 71.5$\pm$0.2 &   15.1$\pm$0.2   & 24.9$\pm$0.5& 30.1$\pm$0.7\\

    CCIC \cite{boschini2022continual} (5120)        & 55.2$\pm$1.4 & 74.3$\pm$1.7 & \textbf{84.7}$\pm$0.9 &              12.0$\pm$0.3& 29.5$\pm$0.4& 44.3$\pm$0.1\\
    ORDisCo \cite{wang2021ordisco} (12500)   & 41.7$\pm$1.2  & 59.9 $\pm$1.4 & 67.6$\pm$1.8  & - & - & -  \\
        CSL (5120)        & 64.3$\pm$0.7 & 73.1$\pm$0.3 & 73.9$\pm$0.1 &   23.7$\pm$0.5         & 41.8$\pm$0.4& 50.3$\pm$0.8\\
    \rowcolor{predcolor} \ourmethod (5120)        & \textbf{73.7}$\pm$0.4 &  \textbf{79.3}$\pm$0.3 &  81.0$\pm$0.2 &  \textbf{27.5}$\pm$0.7 & \textbf{46.0}$\pm$0.2 &  \textbf{56.4}$\pm$0.5 \\
    \bottomrule
  \end{tabular}
  \vspace{-0.2cm}
  \caption{Average accuracy with standard deviation of different methods tested with 5-task CIFAR-10 and 10-task CIFAR-100 settings. The number between brackets indicates the size of the memory buffer for the labeled data.
  }
  \vspace{-0.3cm}

  \label{tab:cifar}
\end{table*}

\subsection{Experimental settings}
\noindent \textbf{Datasets}. We evaluate our method on three datasets. {CIFAR-10} \cite{krizhevsky2009learning} is a dataset of 10 classes with 50k training and 10k testing images. Each image is of size $32 \times 32$. {CIFAR-100} \cite{krizhevsky2009learning} is similar to CIFAR-10, except it has 100 classes containing 500 training images and 100 testing images per class. {ImageNet-100} \cite{tian2020contrastive} is a 100-class subset of the ImageNet-1k dataset from the ImageNet Large Scale Visual Recognition Challenge 2012, containing 1300 training images and 50 test images per class. 

\noindent \textbf{Continual semi-supervised setting}. For both CIFAR-10 and CIFAR-100, we train the models with three different levels of supervision, i.e., $\lambda \in \{0.8\%, 5\%, 25\%\}$. For instance, this corresponds to 4, 25, 125 labeled samples per class in CIFAR-100. As for ImageNet-100, we opt for 1\% labeled data. To build the continual datasets, we use the standard setting in the literature~\cite{wang2021ordisco, boschini2022continual}, and divide the datasets into equally disjoint tasks: 5/10/20 tasks for CIFAR-10/CIFAR-100/ImageNet-100, i.e., 2/10/5 classes per task, respectively. 
We follow the standard class-incremental learning setting in all our experiments: during the CL stages, we assume that all the data of previous tasks are discarded. A memory buffer can be eventually built, but only for labeled data. Following \cite{boschini2022continual}, we set the buffer size for labeled data as 500 or 5120, to ensure a fair comparison. 

\noindent \textbf{Metrics.} We mainly evaluate the performance of different methods considering the average accuracy over all the seen classes after each task, as is common in CL methods \cite{de2021continual}. Analysis with other metrics, such as forward and backward transfer, can be found in the supplementary material.

\noindent\textbf{Implementation details}. As in \cite{boschini2022continual}, we use ResNet18 as our backbone for all the datasets. We adopt the implementation of \cite{assran2021semi}, unless explicitly stated. Specifically, we use the LARS optimizer \cite{you2017large} with a momentum value of 0.9. We set the weight decay as $10^{-5}$, the batch-size as 256. The learning rate is set to 0.4 for CIFAR-10 and 1.2 for CIFAR-100 and ImageNet-100, respectively. We apply 10 epochs warm-up and reduce it with a cosine scheduler. For the two correlated views of the same sample, we generate two large crops and two small crops of each sample. The large crops serve as targets for each other, whereas they are both targets for the small crops. We apply data-augmentation as in \cite{chen2020simple}. Label smoothing is applied with a smoothing factor of 0.1. The additional linear evaluation head is a simple linear layer, which we use to predict labels at test time. % We believe that a more dedicated sampling strategy will help the learning, however, since it is not the main objective of this paper, we leave it for the future work. 
We choose $\lambda_{\operatorname{NND}}=0.2$  and $\lambda_{\operatorname{LIN}}=0.005$. As for the memory buffer, we utilize a simple random sampling strategy. We run our experiments 3 times with different random seeds. The standard deviation is also reported, if applicable. Further implementation details and analysis of data augmentation can be found in the supplementary material.

\noindent \textbf{Baselines.}
 As baselines, we first consider traditional fully-supervised CL methods. A straightforward way to convert them into a continual semi-supervised setting is to use only the labeled data during training and to discard the unlabeled data. We consider two categories of methods: 
 \changes{replay-less methods, namely, continually fine-tuning the model on each new task, online elastic weight consolidation (oEWC) \cite{kirkpatrick2017overcoming}, and replay-based methods, i.e., experience replay (ER) \cite{rolnick2019experience}, iCaRL~\cite{rebuffi2017cvpr}, FOSTER~\cite{wang2022foster} and X-DER~\cite{boschini2022class}.}
 % regularization-based methods, namely, learning without forgetting (LwF)~\cite{li2018pami}, online elastic weight consolidation (oEWC) \cite{kirkpatrick2017overcoming}, and replay-based methods, i.e., experience replay (ER) \cite{rolnick2019experience}, iCaRL~\cite{rebuffi2017cvpr} and GDumb~\cite{prabhu20}. 
 We denote PseudoER as an additional two-stage baseline method, which is a combination of semi-supervised learning (PAWS) and CL (ER) methods. More precisely, we continually train a PAWS and use it to self-label the unlabeled data. An ER method is applied afterward on the labeled and pseudo-labeled data.
We also consider continual semi-supervised learning baseline methods, such as CCIC \cite{boschini2022continual} and ORDisCo \cite{wang2021ordisco}. While CCIC has an explicit definition of the memory buffer for labeled data, which is either 500 or 5120, ORDisCo directly stores all the labeled data that the model receives. We thus denote its memory buffer size with the largest value, which is $M=12500$, equivalent to 25\% of CIFAR-10. As for the upper bound, which is commonly shown in CL~\cite{de2021continual}, it is obtained by jointly training the model with all the available data. \changes{The lower bound refers to continual fine-tuning. CSL is also included as an ablation of \ourmethod  without NND loss.}

\subsection{Results}
\label{subsec:results}

\noindent \textbf{CIFAR-10 and CIFAR-100.} We first report  the performance of different methods on CIFAR-10 and CIFAR-100 in Tab.~\ref{tab:cifar}. The upper bound on CIFAR-10 is  92.1$\pm$0.1\%, and that on CIFAR-100 is  67.7$\pm$0.9\%. \ourmethod outperforms all the competitors in all settings but one, with a significant margin. For instance, when using a buffer of 5120 and 0.8$\%$ of labeled data, \ourmethod performs better than or substantially matches almost all the others, even when they use 25\% labeled data, i.e., about 30 times more supervision. It is also interesting to note that \ourmethod has a very low variance ($\leq 0.7$) across all the settings, indicating a better convergence and representation learning during training. 

From the results in Tab. \ref{tab:cifar}, 
the memory buffer is shown to be effective to alleviate forgetting, as replay-based methods significantly outperform regularization-based ones. 
PseudoER performs better than ER when labeled data is limited (0.8\%), but underperforms when the ratio is higher (5\%, 25\%). We conjecture this as a result of noisy pseudo-labeled data replacing the ground truth in the memory buffer by the sampling strategy of ER, which causes stronger forgetting. In addition, PseudoER is upper-bounded by PAWS, since ER is dependent on the accuracy of pseudo-labeling.

Methods such as CCIC and ORDisCo\footnote{Note that the results of ORDisCo are directly provided by the authors of \cite{wang2021ordisco} as there is no open-source implementation of their approach.} benefit from their design specific to the continual semi-supervised learning scenario. Even though ORDisCo has a larger memory buffer, its performance is inferior to that of CCIC. We believe that this is due to the difficulty in jointly training a continual classifier with a GAN model. CCIC performs reasonably well on CIFAR-10, especially with a large memory buffer. When the buffer size is 5120, CCIC performs better than \ourmethod in the 25$\%$ setting. We suspect that our method underfits in this setting, due to the very small weight of the linear evaluation loss. In contrast, CCIC relies more on labeled data, with equal or even higher importance for the supervised loss. To validate our hypothesis, we increased the weight $\lambda_{\operatorname{LIN}}$ for our linear classifier loss. We obtained an accuracy of 84.5$\pm$0.4\%, which is comparable with CCIC in the same setting (84.7$\pm$0.9\%). 
% It is also worth noting that the 5-task CL setting in CIFAR-10 is a relatively easy problem, wherein we can almost safely discard the unlabeled data if 25\% of data is labeled and the memory buffer is more than 10\% of the size of the dataset.  
In the case of a dataset with more classes such as CIFAR-100, the superiority of \ourmethod is clearly evident. Further comparison with the replay strategy of ORDisCo can be found in the supplementary material.

\begin{figure}[t]
    \centering
    \includegraphics[width=\linewidth]{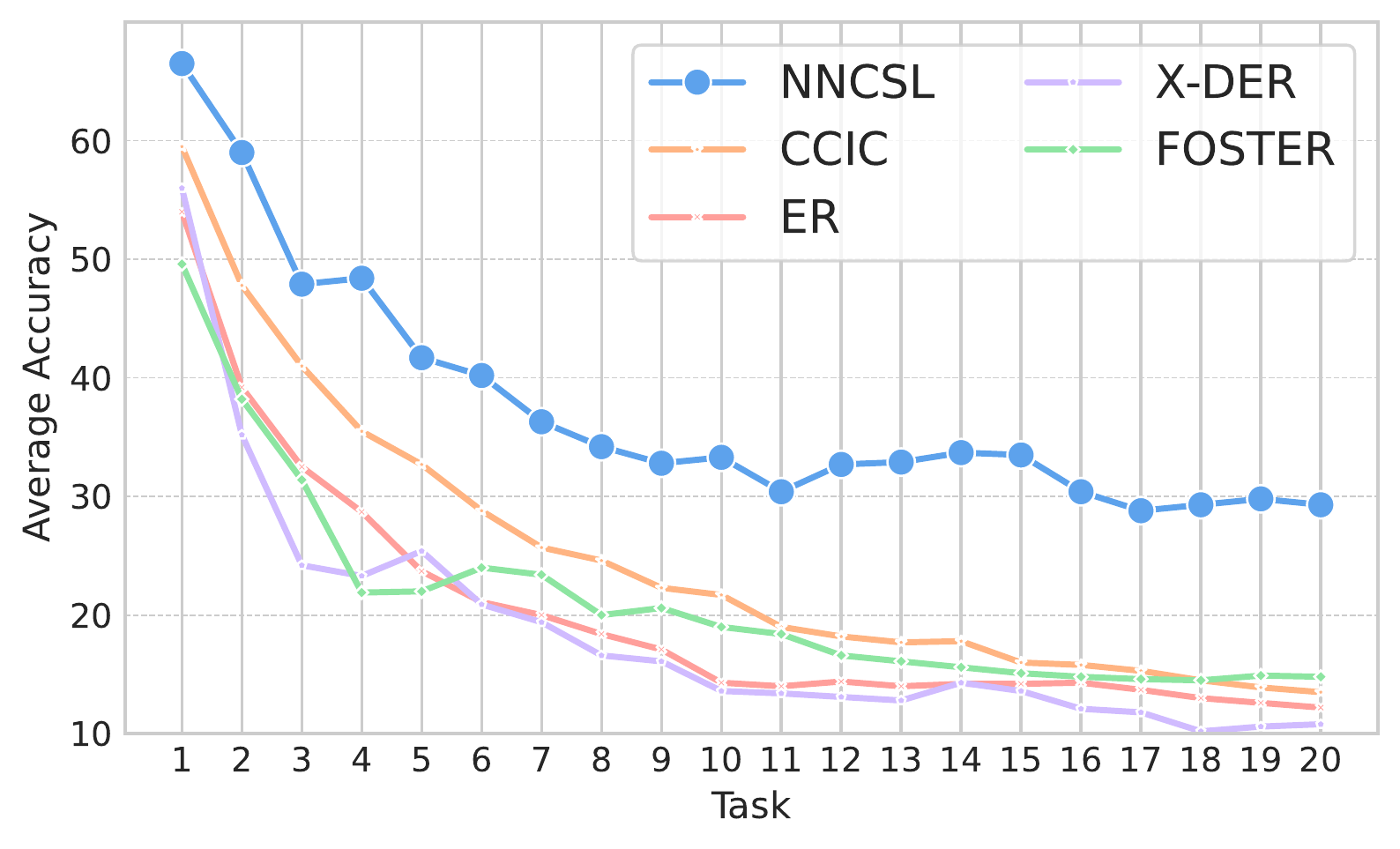}
    \vspace{-0.8cm}
    \caption{Comparison of NNCSL with existing methods on 20-task ImagetNet-100 and 1\% of labeled data. The average accuracy after each learning step is shown.}
    \vspace{-0.25cm}
    \label{fig:20_tasks}
\end{figure}

\noindent \textbf{ImageNet-100.} %ImageNet-100 has never been used by continual semi-supervised learning literature, as it is more challenging than C10 and C100.
\changes{We also evaluate on the more challenging benchmark of ImageNet-100 in a 20-task continual semi-supervised setting and buffer size 5120. As shown in Tab.~\ref{tab:imagenet100_full}, \ourmethod is the best-performing method. CCIC does not show significant improvement with more labeled data (5\% \& 25\%) and also becomes inferior to vanilla CL methods, e.g., ER. We suspect this is a limitation of its representation learning for extracting information from images with higher resolution and larger variance, as we observe its training and validation accuracy in this setting is significantly lower. 
In addition, Fig.~\ref{fig:20_tasks} illustrates the evolution of average accuracy. We note that the average accuracy of our NNCSL stabilizes at around 30\% after the $10^{th}$ task and has a clear rebound between tasks 11 and 16, showing that \ourmethod can effectively retain knowledge acquired during the continual learning steps. In contrast, previously competitive methods clearly suffer from forgetting in this challenging setting, especially in the first several tasks. Despite the same rebounding effect in some methods, e.g., FOSTER, they fail to remain close to NNCSL.}

\begin{table}[t]
  \centering
  \begin{tabular}{l*3c}
  \toprule
    % \textbf{Dataset} & \multicolumn{3}{c}{CIFAR-10(92.12$\pm$0.1)} & \multicolumn{3}{c}{CIFAR-100(67.7$\pm$0.9)}   \\
    \multirow{2}[1]{*}{\textbf{Method}} & \multicolumn{3}{c}{\textbf{ImageNet-100}}   \\
\cmidrule(lr){2-4}
     & 1\% & 5\% & 25\%   \\
    \midrule
    % Joint-training &  &  &  \\
    Fine-tuning &  1.5$\pm$0.2 & 2.7$\pm$0.1 &  4.1$\pm$0.2 \\ 
    ER  \cite{rolnick2019experience} (5120)        & 12.2$\pm$0.8  & 26.3$\pm$0.7 & 38.8$\pm$1.0\\
    % iCaRL  \cite{rebuffi2017cvpr}     &  2.0 &  &  \\
    % \midrule 
     % CCIC \cite{boschini2022continual} (5120)  &  2.9 & 3.4 & 3.8  \\
    FOSTER \cite{wang2022foster} (5120)      & 14.8$\pm$1.1 & 32.8$\pm$0.7 & 42.1$\pm$1.5 \\
    X-DER \cite{boschini2022class} (5120)      & 10.8$\pm$1.1 & 27.4$\pm$1.6 & 45.3$\pm$1.0 \\
    CCIC \cite{boschini2022continual} (5120)       & 13.5$\pm$1.2 & 19.5$\pm$0.7 & 25.9$\pm$0.9   \\
    
    \midrule
    CSL (5120)        & 26.8$\pm$0.4 & 47.9$\pm$0.2 & 56.3$\pm$0.5 \\
    \rowcolor{predcolor} \ourmethod (5120)        & \textbf{29.7}$\pm$0.4 &  \textbf{51.3}$\pm$0.1 & 
    \textbf{65.6}$\pm$0.3\\
    \bottomrule
  \end{tabular}
  \vspace{-0.2cm}
  \caption{Average accuracy with standard deviation of different methods tested with 20-task ImageNet-100 settings. The number between brackets indicates the size of the memory buffer for the labeled data.} 
  \vspace{-0.2cm}
  \label{tab:imagenet100_full}
\end{table}

\subsection{Additional analyses}
\noindent \textbf{Ablation}. We ablate the components of our framework in Tab.~\ref{tab:ablation} on both CIFAR-10 (C10 in the table), CIFAR-100 (C100) and ImageNet100 (IN100). The largest improvement comes from filtering (10.8\% and 6.2\% improvements on CIFAR-10 and CIFAR-100 respectively) and distillation (9.4\% and 4\%  improvements on CIFAR-10 and CIFAR-100 respectively). This confirms the contributions of our proposed components. Although the linear evaluation loss does not bring a significant improvement on the overall performance, we find it useful for stabilizing the learning process, especially for a small dataset with very limited supervision. 
% We justify the necessity of this term in the following analysis. 
Interestingly, PAWS and CSL (w/o filter) diverged on ImageNet-100, as they cannot learn an effective representation with increased scale, due to the instability introduced by distribution drifts and forgetting of unlabeled data (Sec.~\ref{sec:method}). We have included additional analysis on this collapse behavior in the supplementary material.

\begin{table}[t]
  \centering
  \scriptsize
  \begin{tabular}{lcccccc}
  \toprule
  \multirow{2}[1]{*}{\textbf{Method}} & 
\multicolumn{3}{c}{\textbf{Component}} & \multicolumn{3}{c}{\textbf{Dataset}}   \\
\cmidrule(lr){2-4}
\cmidrule(lr){5-7}
    & Distill & Filter & Linear & C10 & C100 & IN100 \\
  \midrule
     PAWS & & & &  51.8&16.1 & Collapse\\
     CSL (w/o filter) & & & \checkmark &   53.0&17.2& Collapse\\
     CSL & & \checkmark &\checkmark &  63.8&23.4&27.1\\
     NNCSL & \checkmark & \checkmark &\checkmark & \textbf{73.2} & \textbf{26.8}&\textbf{29.3}\\
  \bottomrule
  \end{tabular}
  \vspace{-0.2cm}
  \caption{Ablation study of the effectiveness of the proposed components on 5-task CIFAR-10 and 10-task CIFAR-100, with $M=500$ and 0.8\% of labeled data, and 20-task ImageNet-100 with $M=5120$ and 1\% of labeled data. We use average accuracy as metrics.}
  \vspace{-0.15cm}
  \label{tab:ablation}
\end{table}

\vspace{-0.2cm}
\paragraph{Impact of distillation.} Complementary to Fig.~\ref{fig:teaser} (right), which qualitatively shows the superior performance of our proposed NND, we report in Tab.~\ref{tab:distillation} the final accuracy of each task after training on all tasks. An effective distillation method should maintain the performance of old tasks (e.g., task 1 for NND vs.\ feature distillation) and efficiently learn new tasks (e.g., task 5 for NND vs.\ knowledge distillation).

\begin{table}[t]
  \centering
  \small
  \begin{tabular}{lccccc}
  \toprule
    \textbf{Method \& Distillation} & \textbf{T1} & \textbf{T2} & \textbf{T3} & \textbf{T4} & \textbf{T5} \\
    \midrule
        CSL    & 25.2 & 22.9 &  24.6 & 37.3& 37.0 \\
    CSL + Knowledge distill & 27.8 & 24.3 & 22.9 & 35.4 & 31.7 \\
    CSL + Feature distill & 26.5 & 25.8& 26.3& 43.9& 37.1\\
    CSL + NND & \textbf{32.2} & \textbf{26.3} &\textbf{28.1} & \textbf{46.8} & \textbf{38.5}\\
  \bottomrule
  \end{tabular}
  \vspace{-0.25cm}
  \caption{Final accuracy on each task after training on 5-task CIFAR-100. We use CSL as baseline to which we add different distillations. NNCSL is equivalent to CSL + NND.}
  \label{tab:distillation}
  \vspace{-0.2cm}
\end{table}

\begin{figure}
    \centering
\includegraphics[width=\linewidth]{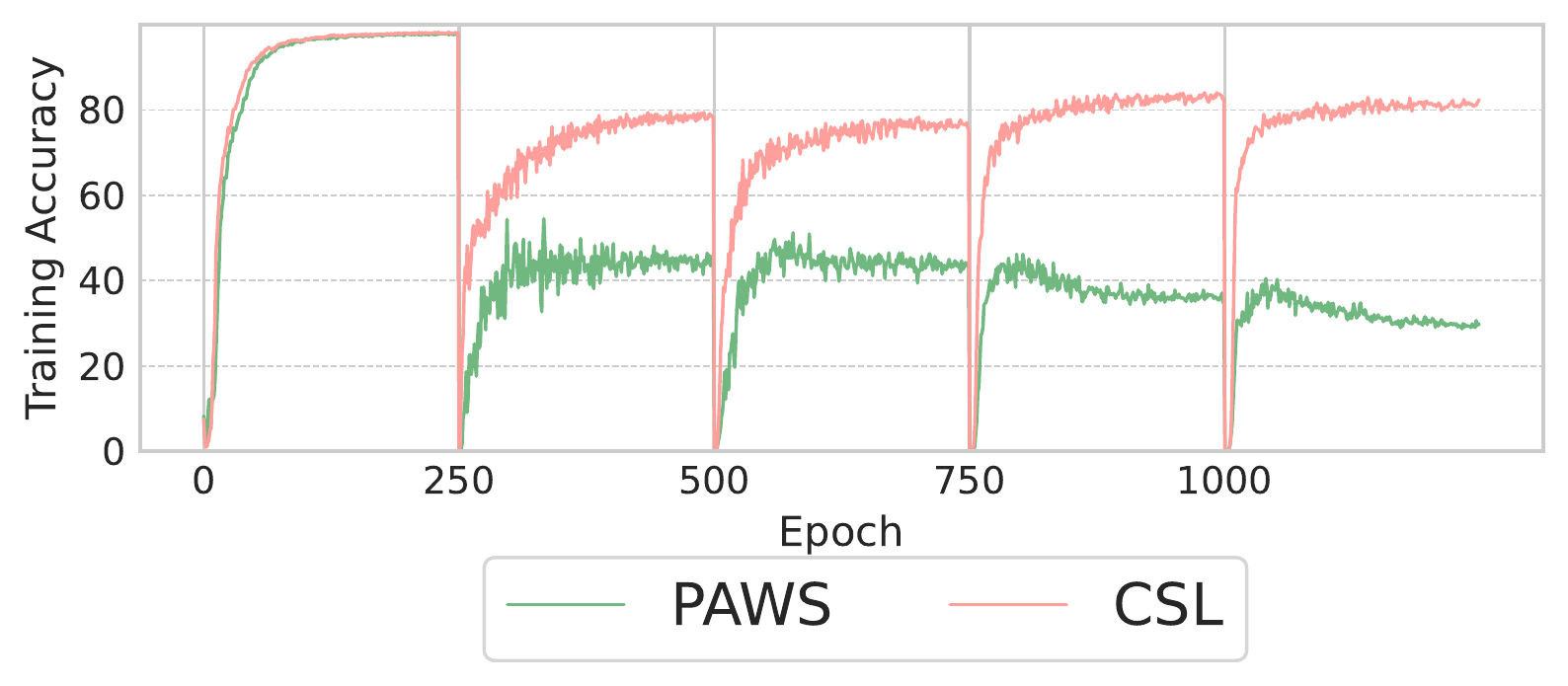}
\vspace{-0.7cm}
\caption{Average training accuracy of unlabeled data on 5-task CIFAR-10. Comparison between vanilla PAWS and our proposed CSL in the continual semi-supervised setting.}
\label{fig:train_acc}
\vspace{-0.2cm}
\end{figure}

%\noindent \textbf
\vspace{-0.3cm}
\paragraph{Evidence of effectiveness.} We visualize in Fig.~\ref{fig:train_acc}  the learning curve of average accuracy for unlabeled training data. 
As we illustrate in Sec.~\ref{subsec:filtering}, the vanilla MEM loss is detrimental to the learning as it forces the model to assign incorrect pseudo-labels to the unlabeled data. Our proposed CSL effectively resolves this issue and allows the model to learn a better representation with the unlabeled data. Note that we only use the labels of unlabeled data to monitor the training, and no label information is leaked at train time.

\begin{table}[t]
  \centering
  \begin{tabular}{lcccc}
  \toprule
    $\lambda_{\operatorname{LIN}}$ & 1 & 0.05 & 0.005 & 0.001 \\
\midrule
    Train (labeled) &99.9 & 99.9 &  97.9 & 96.5 \\
    Train (unlabeled)& 74.7 & 76.0 & \textbf{76.3}   & 75.2  \\
 Validation & 77.1 & 78.4  & \textbf{78.9}   & 77.3 \\
  \bottomrule
  \end{tabular}
  \vspace{-0.2cm}
  \caption{Training and validation accuracies on the current task (i.e., evaluate on task $t$ while training on task $t$) with different values of $\lambda_{\operatorname{LIN}}$. Larger weights increase overfitting on the labeled set and reduce generalization. These experiments are conducted on CIFAR-10 with 5 tasks.}
    \vspace{-0.2cm}
  \label{tab:lambda}
\end{table}

%\noindent \textbf
\vspace{-0.2cm}
\paragraph{Linear evaluation head.} We report the train and validation accuracy in Tab.~\ref{tab:lambda}. Note that for the accuracy at train time, we use the average of accuracy on the current task, i.e., evaluation on task $t$ while training on task $t$.  In general, having a high training accuracy of unlabeled data means the model learns a good representation, which leads to a good validation accuracy. One can observe overfitting  when $\lambda_{\operatorname{LIN}}$ grows. This justifies our choice of i) having a small weight for the linear evaluation head, and ii) choosing PAWS as the basis, which does not directly use labeled data as training samples to avoid overfitting. Moreover, we observe underfitting when $\lambda_{\operatorname{LIN}}$ is smaller than 0.005. It confirms the need for this linear classifier in the continual semi-supervised scenario. We thus set $\lambda_{\operatorname{LIN}}$ as $0.005$ in our experiments.

\vspace{-0.2cm}
\paragraph{Ablation of memory buffer.} To verify the effectiveness of the replay buffer, we evaluate our \ourmethod without this and observe a drastic decrease in performance (19.7\% vs 73.2\%) on 5-task CIFAR-10 with 0.8\% of labeled data. However, our method recovers 95\% of the performance (69.2\% vs.\ 73.2\%) while using only 10\% of memory (50 vs.\ 500). We conjecture such data efficiency results from the linear classifier that provides strong gradients in our method. \changes{More details of this study can be found in the supplementary material.}

\section{Conclusion}
\vspace{-0.1cm}
\label{sec:conclusion}
In this work, we studied continual semi-supervised learning and proposed \ourmethod, a novel approach based on soft nearest-neighbors and distillation. Our extensive experiments show the superior performance of \ourmethod with respect to existing methods, setting a new state of the art.
Previous work~\cite{assran2021semi} showed that using a more powerful network such as wider or deeper ResNet can further improve performance. While this is not addressed in this work, we consider it an interesting direction for future work. In this paper, we considered a fixed ratio between labeled and unlabeled samples across all tasks. A varying ratio would be an even more challenging setting for future investigation.
\vspace{-0.4cm}
\paragraph{Acknowledgements} This work was funded in part by the ANR grant AVENUE (ANR-18-CE23-0011). It was also granted access to the HPC resources of IDRIS under the allocation 2021-[AD011013084] made by GENCI.

\singleappendix{Appendix}
\appendix
\label{sec:appendix}
\section{Implementation details}
Although our model shares most of the hyper-parameters across different datasets, there are few differences, in values chosen empirically, to adapt to different scenarios. NNCSL, as well as PAWS~\cite{assran2021semi} and CSL for ablation study, are trained with 250 epochs per task for CIFAR-10 and CIFAR-100, and 100 epochs for ImageNet-100. For CIFAR-10, the learning rate is initialized as 0.08, warmed up to 0.4, and reduced to 0.032 with the cosine scheduler. For CIFAR-100, a similar variation of learning rate is set from 0.08 to 1.2 to 0.032, and for ImageNet-100, it is 0.3 to 1.2 to 0.064. The color distortion ratio is set to 1 for ImageNet-100 and 0.5 for CIFAR-10 and CIFAR-100. The size of the mini-batch for labeled data is set to 5 for CIFAR-10 and 3 for CIFAR-100 and ImageNet-100. The size of the mini-batch for unlabeled data is set to 256 for CIFAR-10 and CIFAR-100 and 64 for ImageNet-100. These hyper-parameters are mostly based on the suggested default values of PAWS, and we empirically update them after testing with a moderate set of values variant around the default ones, based on the validation performance. However, We do not perform hyper-parameter tuning on ImageNet-100: we first adopt the hyper-parameters for ImageNet from PAWS and update them with the same changes we apply on CIFAR-100.

For the continual learning setting, we initialize a unified linear evaluation head where the number of outputs is the total number of classes in the dataset. When a class is not yet seen by the model, the corresponding output is masked. To retain a copy of the previously trained model, we use the \textit{deepcopy} method from the \textit{copy} package\footnote{https://docs.python.org/3/library/copy.html} 

The copied model is in evaluation mode when training the current model on the new classes.

We have included our source code as part of the supplementary material. All the  implementation details can be found in the options files, for instance, random seeds, and labeled samples on each dataset.  We plan to open-source our code upon acceptance of this submission.

\section{Comparison of data augmentation}
We note that the data augmentation of our proposed framework is not the same as the one used in CCIC \cite{boschini2022continual}. CCIC utilizes random cropping and horizontal flipping (which we refer to as \textit{weak DA}), whereas our proposed CSL and NNCSL include color distortion as an additional operation for data augmentation (referred to as \textit{strong DA}). To verify the impact of this additional augmentation strategy, we include color distortion in the data augmentation process of CCIC and re-train it from scratch on CIFAR-10 (C10 in the table) with 5 tasks, 5\% labeled data and buffer size 5120, and also on CIFAR-100 (C100 in the table) with 10 tasks, 0.8\% labeled data and buffer size 5120. The results are reported in Tab.~\ref{tab:data_aug}. CCIC does not benefit from color distortion on both datasets. We believe this is because CCIC does not have the multiple-view consistency to be robust with respect to strong data augmentation. Consequently, we chose to report results using CCIC's original (and more effective) data augmentation.

\begin{table}[t]
  \centering
  \small
  \begin{tabular}{lccc}
  \toprule
\multirow{2}[1]{*}{\textbf{Method}} &  \multirow{2}[1]{*}{\textbf{Dataset}} & 
 \multicolumn{2}{c}{\textbf{Data Augmentation}}    \\
\cmidrule(lr){3-4}
    &   & Weak & Strong \\
  \midrule
     CCIC& C10 & \textbf{72.8} & 69.4  \\
     CCIC & C100 &\textbf{12.0} & 9.9  \\
  \bottomrule
  \end{tabular}
  \caption{Comparison of different data augmentation strategies for CCIC on CIFAR-10 (denoted as C10 in the table) and CIFAR-100 (denoted as C100 in the table).}
  \label{tab:data_aug}
\end{table}

\begin{table}[t]
  \centering
  \begin{tabular}{lcc}
  \toprule
     \textbf{Method} & \textbf{Replay strategy} & \textbf{Average Accuracy} \\
    \midrule
    NNCSL & Labeled & \textbf{76.7} \\
    NNCSL & Labeled \& Unlabeled & \textbf{82.1} \\
    ORDisCo & Labeled \& GR & 65.9 \\
  \bottomrule
  \end{tabular}
  \caption{Comparison of different strategies for the replay buffer with 5-task CIFAR-10, using 3\% of labeled data to match the setting of \cite{wang2021ordisco}.}
  \label{tab:unlabel_strat}
\end{table}

\section{Replay strategies}
ORDisCo~\cite{wang2021ordisco} utilizes a generative replay (GR) strategy to replay unlabeled data. Given that the generative model brings a memory overhead that is not negligible, it is reasonable to equip our method with a memory buffer for unlabeled data for a fair comparison. Specifically, we use 5000 samples, which is equivalent to the size of the generative model of ORDisCo. Tab.~\ref{tab:unlabel_strat} shows that having access to the previously seen unlabeled data can indeed improve the performance of our method, and our NNCSL performs better with a simple memory buffer than ORDisCo with a sophisticated generative-replay strategy. This experiment confirms the ability of our method to exploit unlabeled data.

\begin{table}[t]
  \centering
  \small
  \begin{tabular}{lcccccc}
  \toprule
    Buffer size & 0 & 8 & 16 & 50  & 500 & 5120 \\
\midrule
    NNCSL & 19.7 & 37.7 &  53.2 & 69.2 &   73.2 & 73.7\\
  \bottomrule
  \end{tabular}
  \vspace{-0.2cm}
  \caption{CIFAR-10 average accuracy wrt memory buffer sizes with 5 tasks, 0.8\% of labeled data.}
  \label{tab:buffer_size}
  \vspace{-0.2cm}
\end{table}

\section{Ablation study of the memory buffer}
We present the ablation study of the memory buffer in a table, as shown in Tab.~\ref{tab:buffer_size}, for better readability. The use of a memory buffer is a common practice in CL and most replay-based methods would fail if the buffer is removed. For instance, we observe a drastic decrease in performance (19.7\% vs. 73.2\%) when no replay buffer is allowed. In addition, increasing the memory buffer leads to improvements in performance with a certain upper bound. The performance plateaus when the memory buffer is larger than 500 (500 vs. 5000) because the buffer size is larger than the total number of labeled data in this case (i.e., 400 labeled samples).

\section{Ambiguity of two-stage methods in CL}

\begin{figure}
    \centering
    \includegraphics[width=\linewidth]{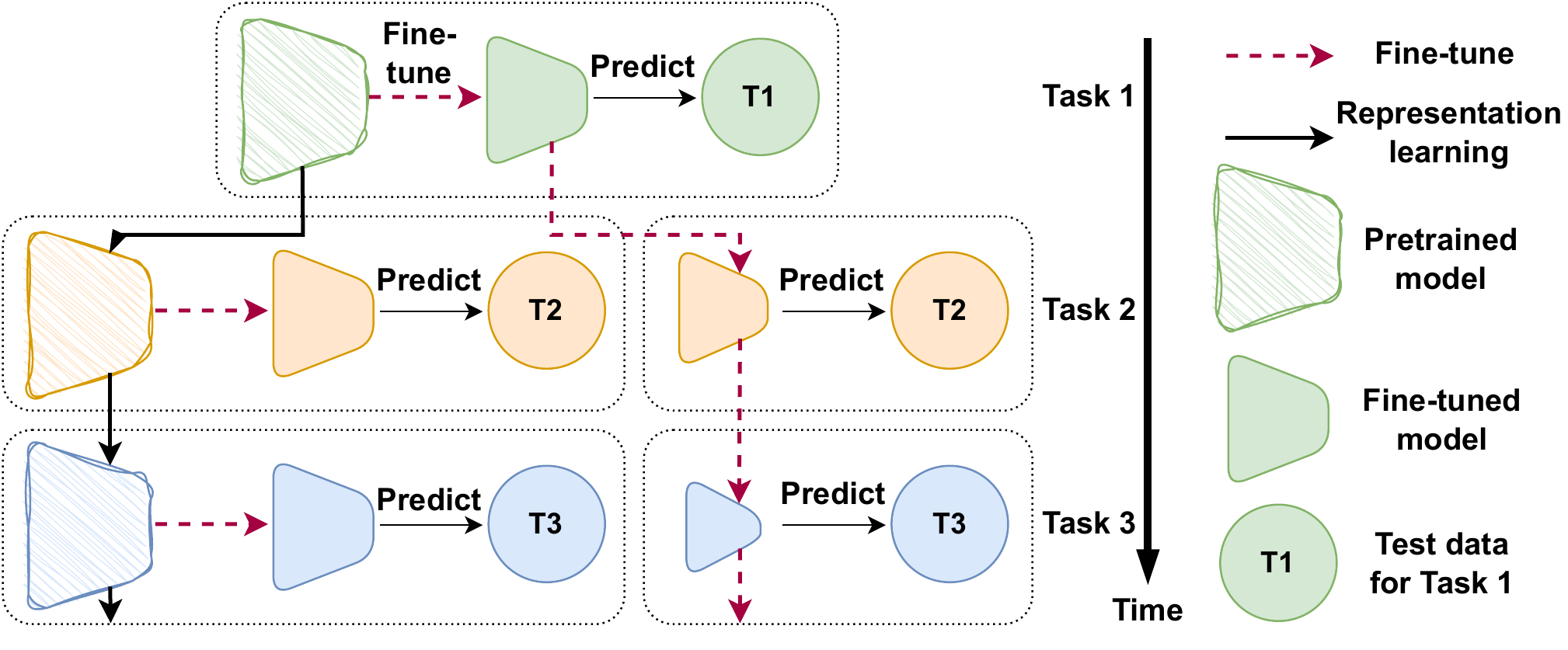}
    \vspace{-0.7cm}
    \caption{\changes{The ambiguity of two-stage methods in CL. On the right path, the shrinking size of the fine-tuned model after each task shows the loss of generalizability.}}
    \label{fig:two_stage}
    \vspace{-0.3cm}
\end{figure}

\changes{We show in Fig.~\ref{fig:two_stage} how two-stage methods (pre-training and then fine-tuning) can be adapted to a CL setting. For instance, after training on Task 1, one has to choose the model to be utilized in the subsequent tasks. The left path reuses the pre-trained model. It ensures the generalizability for learning each new task but results in additional memory overhead, as the fine-tuned model has to be saved for testing. In contrast, the right path reuses the fine-tuned model, which leads to a unified model for all tasks. However, the overfitting caused by the small labeled set is detrimental to the generalizability of the model. We explicitly show such loss in Fig.\ref{fig:two_stage} by associating the size of the fine-tuned model with its generalizability. On the right path, the fine-tuned model is shrinking due to the overfitting issue.}

\section{Training analysis on ImageNet-100}
\begin{figure}
\begin{subfigure}[h]{\linewidth}
\includegraphics[width=\linewidth]{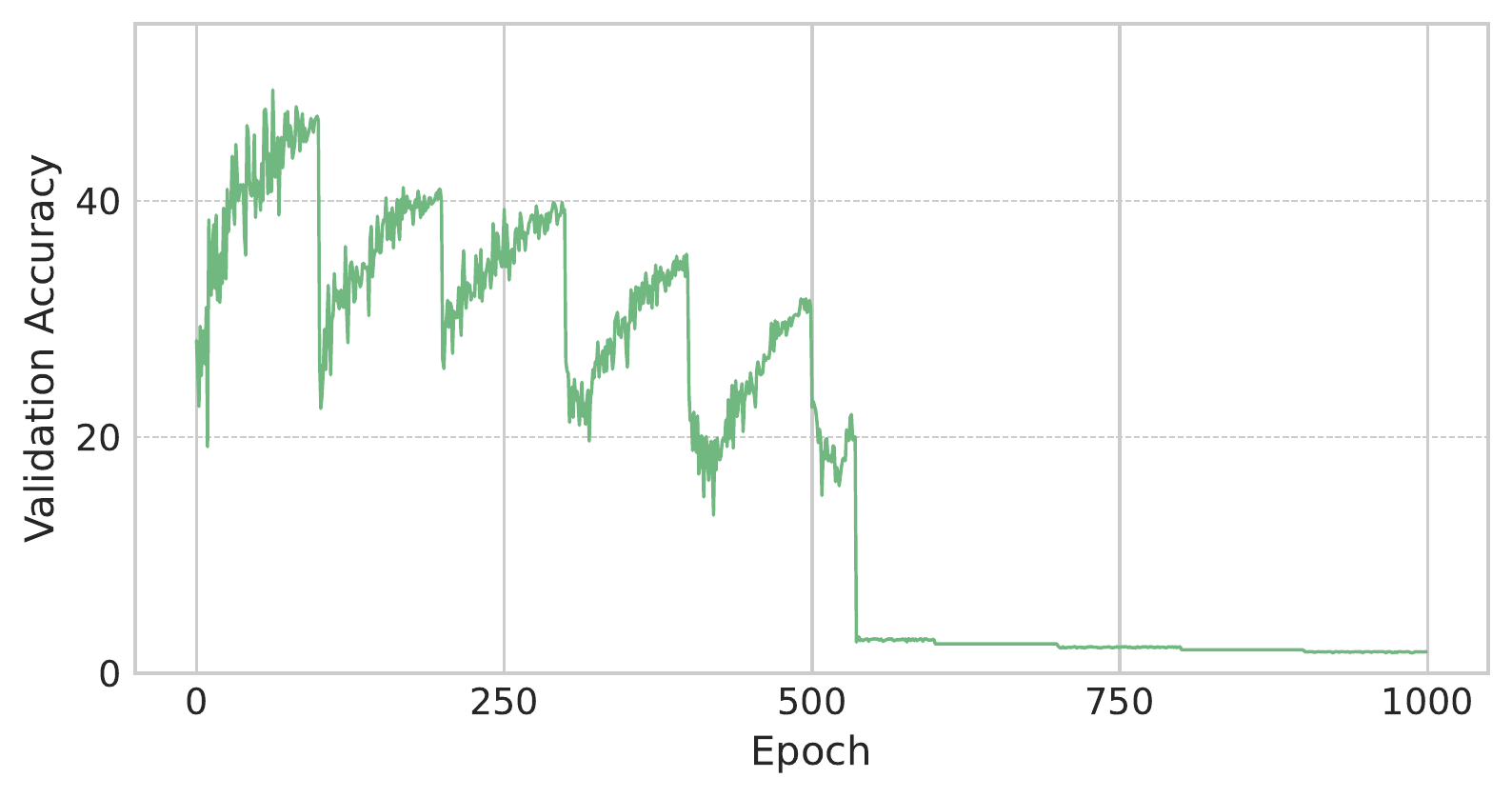}
\caption{Validation accuracy of PAWS}
\end{subfigure}
\begin{subfigure}[h]{\linewidth}
\includegraphics[width=\linewidth]{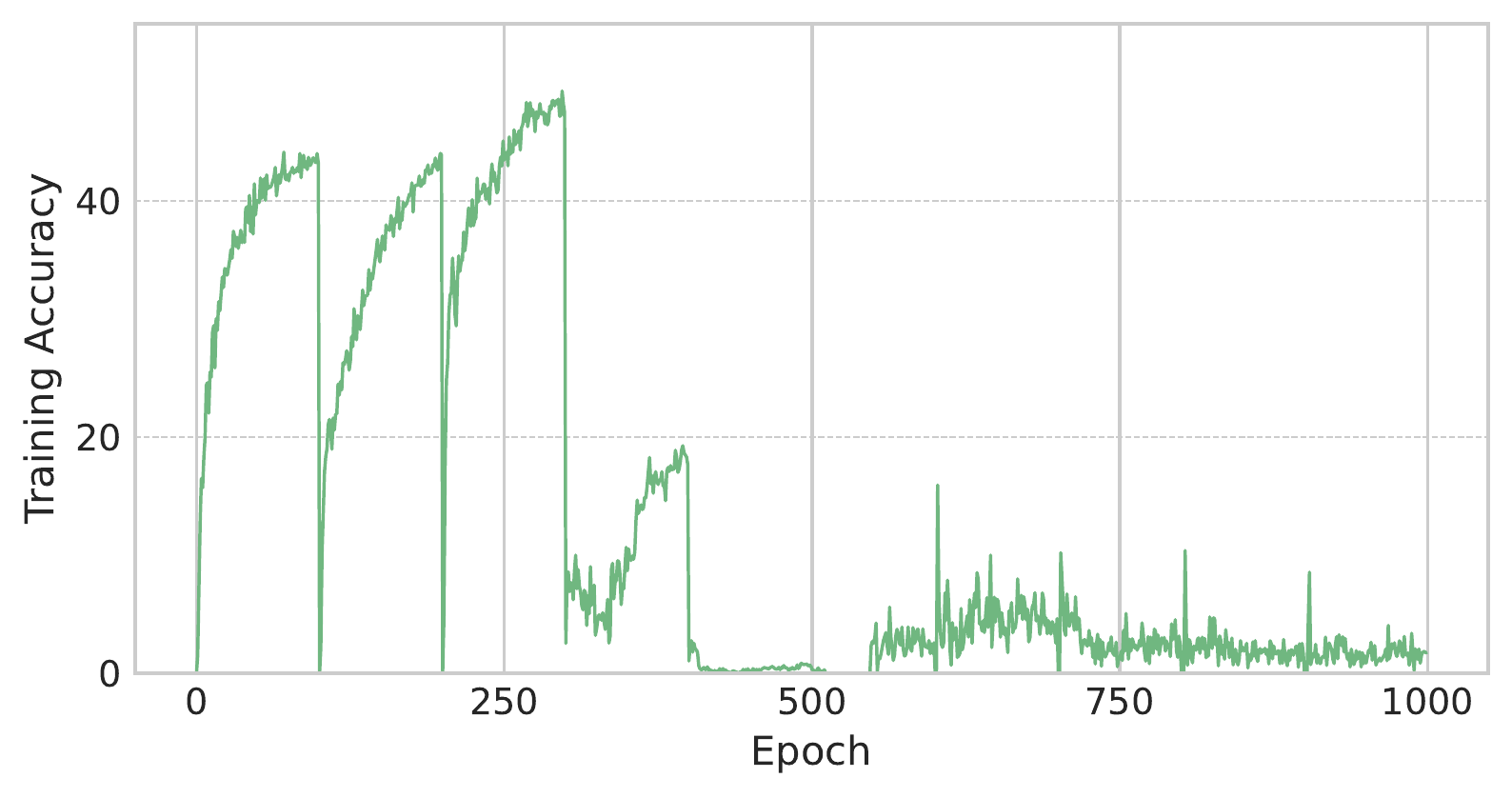}
\caption{Training accuracy for unlabeled data of PAWS}
\end{subfigure}
\caption{Learning curve of PAWS on ImageNet-100.}
\label{fig:imgnt100_paws}
\end{figure}

It is interesting to see that PAWS diverges in this setting, as is shown in the Tab. 3 of the main paper. Our analysis reveals that the vanilla MEM loss strongly impacts representation learning on unlabeled data and makes the learning procedure highly unstable, as shown in Fig.~\ref{fig:imgnt100_paws}. Although we do not observe the same collapse of PAWS on CIFAR-10 or CIFAR-100, recall that in Fig. 6
% .~\ref{fig:train_acc} 
of the main paper, the training accuracy of unlabeled data for PAWS is strongly constrained on CIFAR-10. This means the representation learning of PAWS is already vulnerable. As images of ImageNet-100 have a much larger resolution than that of CIFAR-10, learning a robust feature from the input of ImageNet-100 is significantly more difficult. In such a complex case, the model easily diverges but can hardly recover, we suspect that it is because the gradient is very noisy (due to the negative impact of MEM loss) and small (due to the partial supervision and indirect use of labeled data). To verify this assumption, we observe that adding the linear head slightly alleviates the divergence. However, it cannot prevent the collapse from happening. This means that the MEM loss is the main cause of the collapse and is indeed detrimental to representation learning.

Nevertheless, we believe it may be possible to resolve this collapse without changing the framework, i.e., PAWS. For example, one can conduct careful, extensive hyper-parameter tuning to find an optimal set of parameters that can stabilize the learning. However, it is not realistic given the scale of the dataset. Hence, we did not conduct such experiments.

\section{Impact of $\lambda_{\operatorname{NND}}$}
In Tab.~\ref{tab:lambda_NND}, we report the performance of our NNCSL with respect to different values of $\lambda_{\operatorname{NND}}$ on CIFAR-100, with 5 tasks, 1\% of labeled data and buffer size 5120. $\lambda_{\operatorname{NND}}$ controls the importance of the distillation branch with respect to the PAWS loss. The higher the value, the stronger constraint the model receives to retain the previous knowledge. $\lambda_{\operatorname{NND}}= 0$ means no distillation, which reduces the model back to CSL. We can clearly see that distillation helps the model perform better (e.g., $\lambda_{\operatorname{NND}}= 0$ vs. $\lambda_{\operatorname{NND}}= 0.2$) and too much regularization from distillation can constraint the model from learning new knowledge (e.g., $\lambda_{\operatorname{NND}}= 0.2$ vs. $\lambda_{\operatorname{NND}}= 1$). 
\begin{table}[t]
  \centering
  \begin{tabular}{lccccc}
  \toprule
    $\lambda_{\operatorname{NND}}$ & 0 & 0.01 & 0.1 & 0.2  & 1 \\
\midrule
    NNCSL & 29.0 & 30.2 &  31.8 & \textbf{33.6} &   30.5\\
  \bottomrule
  \end{tabular}
  \caption{Average Accuracy with different values of $\lambda_{\operatorname{NND}}$. These experiments are conducted on CIFAR-100 with 5 tasks, 1\% of labeled data and buffer size 5120.}
  \label{tab:lambda_NND}
\end{table}

\section{Forward and backward transfer analysis}
Forward transfer (FWT) and backward transfer (BWT) are commonly used in continual learning literature \cite{rebuffi2017cvpr, Lopez-Paz17}. The former measures the capacity of the model to generalize to future tasks, whereas the latter shows the capacity of the model to retain the previously acquired knowledge. Specifically, they are defined as follows. Let $T$ again be the total number of tasks for the continual learning stages, we therefore can divide the test set into $T$ segments, each one representing one task. After each task $t$, the model is evaluated with respect to all $T$ test sets. Consequently, we obtain a matrix $R \in \mathbb{R}^{T \times T}$, where the element $R_{i,j}$ is the test performance on task $j$ with the model on task $i$. We use \textit{classification accuracy} as our evaluation metrics. In addition, we define the random estimation as $r_j$, which represents the test performance on task $j$ using a model with only random initialization. We can define the FWT and BWT as:

\begin{equation}
    FWT =  \frac{1}{T-1} \left(\sum^{T}_{i=2} R_{i-1, i} - r_{i}\right).
\label{eq:fwt}
\end{equation}

\begin{equation}
    BWT =  \frac{1}{T-1} \left(\sum^{T-1}_{i=1} R_{T, i} - R_{i,i}\right).
\label{eq:bwt}
\end{equation}

\noindent Similarly, we can define the average accuracy (ACC) as:

\begin{equation}
    ACC =  \frac{1}{T} \left(\sum^{T}_{i=1} R_{T, i}\right).
\label{eq:acc}
\end{equation}

\noindent It should be noticed that computing the backward transfer for the first task or the forward transfer for the last task have little utility and are excluded from Eq.~\ref{eq:fwt} and Eq.~\ref{eq:bwt}. 

We report the results in Tab.~\ref{tab:fwt_bwt} a comparison of our proposed components. Note that PAWS diverges in this setting, leading to a low FWT. Instead, PAWS is better than CSL and NNCSL if we look at BWT alone. It is simply because $R_{T, i}$ and $R_{i,i}$ are all low after the divergence, having not much room for the model to forget. That is, a model cannot forget if it does not learn anything first. This observation confirms the limitation of BWT, as it only shows a relative difference with respect to its own performance, i.e., Eq.~\ref{eq:bwt}. Thus, BWT is more suitable to be an additional indicator when the average accuracy of the two methods is close to each other, e.g., NNCSL vs. CSL. Comparing NNCSL and CSL, we notice that the NND helps slightly improve the BWT. What is more interesting is that NND significantly improves FWT. We believe it is because NND stabilizes the representation learning, allowing the model to generalize better to future tasks.

We also notice that the absolute value of BWT is high for both NNCSL and CSL. We suggest that it is because the first task suffers the most from forgetting, as it is trained with a simple task and without any regularization of distillation, but it goes through all continual stages. To verify this assumption, we compute the BWT without the first task: $-11.3$ for NNCSL and $-9.23$ for CSL, which are significantly improved from the BWT scores in Tab.~\ref{tab:fwt_bwt}.

\begin{table}[t]
  \centering
  \begin{tabular}{ccc}
  \toprule
     \textbf{Method} & \textbf{FWT} $\uparrow$ & \textbf{BWT} $\uparrow$\\
    \midrule
    PAWS  &  1.1 & -13.7  \\
    CSL  & 26.8 &  -18.25 \\
    NNCSL  &  \textbf{31.7}&  \textbf{-17.15}\\
    % Average Forgetting $\downarrow$ & 17.2  & \textbf{17.1}    \\
  \bottomrule
  \end{tabular}
\caption{Forward transfer (FWT) and backward transfer (BWT) of PAWS, CSL and NNCSL in 20-task ImageNet-100. }
\label{tab:fwt_bwt}
\end{table}

\section{Visualization of the features}
We use t-SNE \cite{van2008visualizing} to project the learned features into a lower-dimensional space and visualize them to qualitatively verify the effectiveness of our proposed method. %More precisely, t-SNE is a technique for non-linear dimensionality reduction whose property is that similar samples will be closed to each other whereas dissimilar samples will be pushed away.
We apply t-SNE on the deep features  $\rvh_u = h(\rvz_u)$ of \textbf{unlabeled data} and color them in the visualization with their ground-truth label. Ideally, if the features are well learned, one can see different clusters representing different classes in the visualization. Specifically, we choose 5-task CIFAR-10 to ensure a distinguishable class boundary.

\begin{figure}
    \centering
\includegraphics[width=\columnwidth]{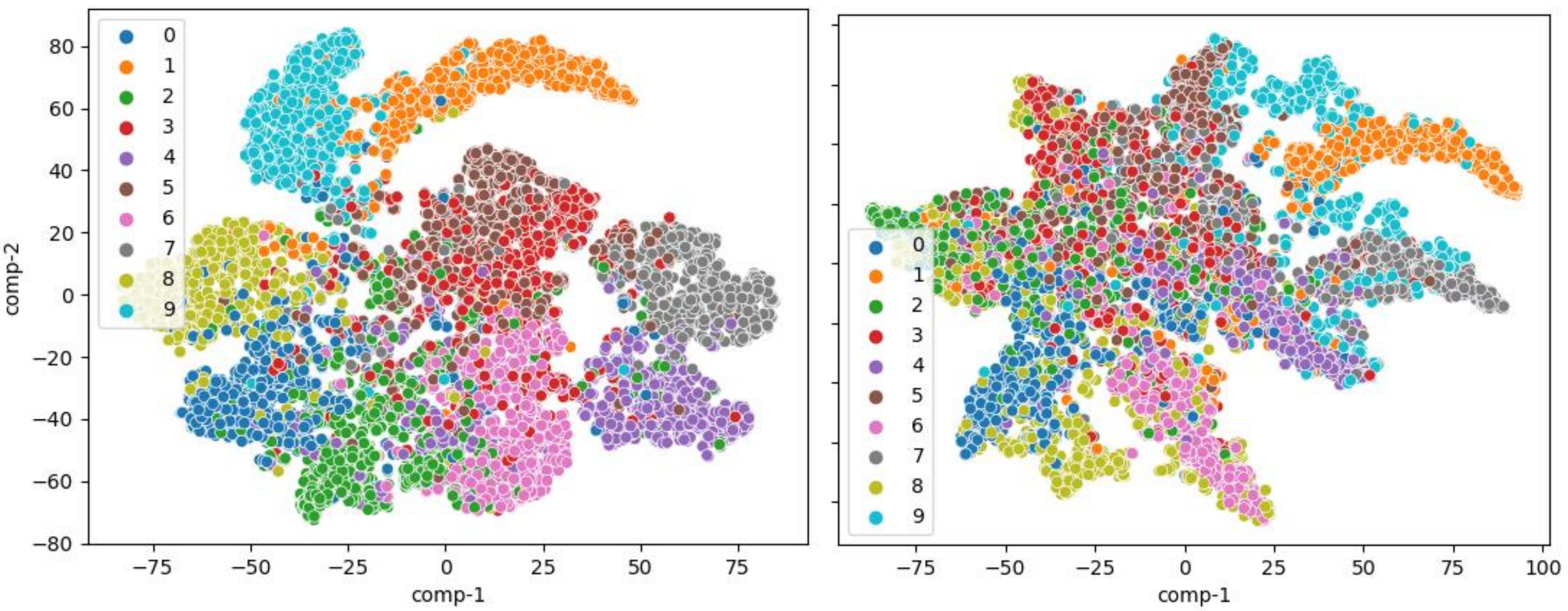}
\caption{T-SNE visualization of deep features of 10 classes of CIFAR-10, these experiments are conducted with 5 tasks. Left: features from NNCSL after training on task 5, Right: features from PAWS after training on task 5. Data points are colored by their corresponding classes. A clear class boundary after several tasks shows a robust representation along the continual learning stages.
}
\label{fig:tsne_10cls}
% \vspace{2.5cm}
\end{figure}
\begin{figure}
    \centering
\includegraphics[width=\columnwidth]{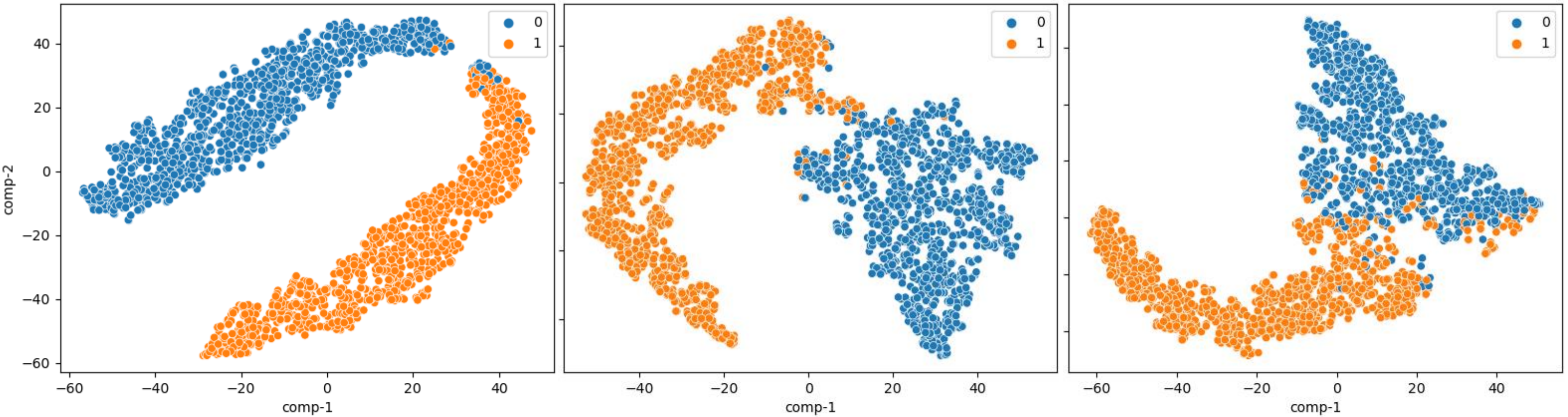}
\caption{T-SNE visualization of deep features of the first 2 classes of CIFAR-10, these experiments are conducted with 5 tasks.  Left: features from NNCSL after training on task 1, Middle: features from NNCSL after training on task 5, Right: features from PAWS after training on task 5. Data points are colored by their corresponding classes. It is clear that PAWS suffers from a blurry class boundary after several continual learning stages. 
}
\label{fig:tsne}
\end{figure}

The result is shown in Fig.~\ref{fig:tsne_10cls}. The figure on the left shows the features of all 10 classes after task 5, using NNCSL. Recall that CIFAR-10 is divided into 5 tasks. We can see a clear separation of different classes in the visualization. Fig.~\ref{fig:tsne_10cls} Right shows the features of the same 10 classes after task 5 using PAWS. We can clearly see that the vanilla MEM loss of PAWS causes a blurry class boundary as it tried to scatter over all classes with partially available unlabeled data. 

To have a more detailed view on the feature space, we select the first two classes as examples and visualize them at different training stages using different methods. Fig.~\ref{fig:tsne} confirms that PAWS leads to a blurry boundary and is prone to severe forgetting due to this effect.

%%%%%%%%% REFERENCES
{\small
\bibliographystyle{ieee_fullname}
\bibliography{biblio}
}

\end{document}